\documentclass[sn-mathphys,Numbered]{sn-jnl}%

\usepackage{geometry}
\usepackage{graphicx}%
\usepackage{multirow}%
\usepackage{amsmath,amssymb,amsfonts}%
\usepackage{amsthm}%
\usepackage{mathrsfs}%
\usepackage[title]{appendix}%
\usepackage{xcolor}%
\usepackage{textcomp}%
\usepackage{manyfoot}%
\usepackage{booktabs}%
\usepackage{algorithm}%
\usepackage{algorithmicx}%
\usepackage{algpseudocode}%
\usepackage{listings}%

\usepackage{array}
\usepackage{caption}
\usepackage{subcaption}

\theoremstyle{thmstyleone}%

\theoremstyle{thmstyletwo}%

\theoremstyle{thmstylethree}%

\newcommand{\figref}[1]{Fig.~\ref{#1}}
\newcommand{\tbref}[1]{Table~\ref{#1}}
\newcommand{\secref}[1]{\S\ref{#1}}

\newcommand{\LMdebates}{\textsc{LM}$_{\text{\textsc{debates}}}$}
\newcommand{\LMsotu}{\textsc{LM}$_{\text{\textsc{sotu}}}$}
\newcommand{\LMcampaign}{\textsc{LM}$_{\text{\textsc{campaign}}}$}

\newenvironment*{revision}{}{}  %

\newenvironment*{revisionTwo}{}{}  %

\newenvironment*{revisionThree}{}{}  %

\begin{document}

\title[Quantifying the Uniqueness and Divisiveness of Presidential Discourse]{Quantifying the Uniqueness and Divisiveness of Presidential Discourse}

\author[a]{\fnm{Karen} \sur{Zhou}}
\author[a,c]{\fnm{Alexander A.} \sur{Meitus}} 
\author[a]{\fnm{Milo} \sur{Chase}}
\author[a]{\fnm{Grace} \sur{Wang}}
\author[d]{\fnm{Anne} \sur{Mykland}}
\author[b,c]{\fnm{William} \sur{Howell}}
\author*[a,c]{\fnm{Chenhao} \sur{Tan}}\email{chenhao@uchicago.edu}

\affil*[a]{Department of Computer Science, The University of Chicago}
\affil[b]{Department of Political Science, The University of Chicago}
\affil[c]{Harris School of Public Policy, The University of Chicago}
\affil[d]{Harvard University}

\abstract{
    Do American presidents speak discernibly different from each other? If so, in what ways? And are these differences confined to any single medium of communication? To investigate these questions, this paper introduces a novel metric of uniqueness based on large language models, develops a new lexicon for divisive speech, and presents a framework for assessing the distinctive ways in which presidents speak about their political opponents. Applying these tools to a variety of corpora of presidential speeches, we find considerable evidence that Donald Trump's speech patterns diverge from those of all major party nominees for the presidency in recent history. Trump is significantly more distinctive than his fellow Republicans, whose uniqueness values appear closer to those of the Democrats. Contributing to these differences is Trump's employment of divisive and antagonistic language, particularly when targeting his political opponents. These differences hold across a variety of measurement strategies, arise on both the campaign trail and in official presidential addresses, and do not appear to be an artifact of secular changes in presidential communications.
}


\maketitle

\begin{center}
    \noindent
    \setlength{\fboxsep}{8pt}
    \setlength{\fboxrule}{1pt}
    \fbox{%
        \parbox[t]{0.95\linewidth}{\vspace{-\baselineskip}%
            \paragraph{Significance Statement}
            {\revisionThree While presidential discourse attracts considerable attention from popular media and scholars alike, efforts to computationally compare such rhetoric have been limited to lexical methods.}
            This paper proposes a novel suite of metrics to advance such analyses and identify new findings. In particular, we leverage large language models to establish an original metric of uniqueness, develop a new lexicon for divisive speech, and introduce a comparative framework for the portrayal of political opponents. 
            {\revision We then apply these methods to a rich assembly of campaign speeches, presidential debates, and official States of the Union addresses. Across all these {\revisionTwo datasets},
            we find that {\revisionThree Donald} Trump's political rhetoric is unique among modern presidents, and is defined, in part, by his use of antagonistic language, particularly when directed at political opponents.}
        }
    }
    \end{center}

\section{Introduction}

{The rise of the modern presidency is defined, in no small measure, by the chief executive's changing relationship to the American public. In what Jeffrey Tulis \cite{tulis1987} calls the ``rhetorical presidency,'' modern presidents are expected to routinely stand before the public in order to explain, persuade, inform, and instruct. How precisely they do so, though, is open to interpretation. While certain norms of communication govern their behavior, presidents have a fair measure of discretion to speak as they choose.}

{When fulfilling their oratory duties, do modern presidents adhere to a common script? Or do some presidents defy rhetorical conventions and speak in ways that, at least among themselves, are novel and surprising? Leveraging recent advances in large language models (LLMs)~\cite{radford2018improving}, we develop a new quantifier of uniqueness by directly measuring {\revision the unpredictability of language patterns.}
In addition, we {\revision incorporate more standard lexical techniques to examine a} prevalent yet understudied construct, divisiveness~\cite{cinar2020}.
We operationalize {\em divisiveness} as language that is intended to impugn and delegitimize the speaker's target, and we develop a new lexicon for such speech. 
Furthermore, we {\revision introduce a comparative framework for assessing mentions of opponents, which are especially prevalent in presidential debates.} 
}

{We use our proposed tools to 
analyze large and diverse corpora of presidential speech. In doing so, we are able to both characterize the overall distinctiveness of presidents' speech patterns and examine specific qualities that previous research has overlooked. Moreover, we can distinguish general speech patterns from those directed towards one's political opponents.} 

In nearly all of our analyses, Donald Trump appears as a clear outlier. On the campaign trail, in presidential debates, and in official presidential addresses, we find, Trump's speech patterns routinely differ from those of all recent presidents {--- lending credence to Kurt Anderson's observation that, ``The version of English [Trump] speaks amounts to its own patois, with a special vocabulary and syntax and psychological substrate'' \cite{anderson2018}.}

Like previous scholars \cite{kayam_readability_2018, Frischling_2019, doi:10.1073/pnas.1811987116}, we find that Trump tends to communicate in shorter, more simplistic sentences. But whereas previous quantitative research on presidential rhetoric relied exclusively on lexicons (e.g., \cite{doi:10.1073/pnas.1811987116}) and sentiment classification (e.g., \cite{Zhong2016TheCI}), which necessarily disregard contextual information included in the body of speeches, our methods are able to show that Trump speaks in ways that are holistically different from all modern presidents. These differences are pervasive and large --- so much so, in fact, that the observed differences between Trump and his fellow Republicans exceed those between Republicans and Democratic presidents. Contributing to these differences, we show, is Trump's tendency to speak in ways that are especially divisive, particularly when focusing on his political opponents.\footnote{This finding is consistent with previous research showing that Trump uses language that is explicitly intended to evoke an emotional response from his revision audience \cite{hart2020}, and that Trump's speech patterns reveal his autocratic ambitions \cite{korner}, anti-democratic views \cite{jamieson2017}, a willingness to condone political violence \cite{ntontis}, a commitment to populism \cite{cinar2020}, and low levels of analytic thinking~\cite{doi:10.1073/pnas.1811987116}. None of these scholars, however, have developed a lexicon of divisiveness, as we do in this paper.} These findings, moreover, are robust to a variety of measurement strategies, arise across rich and diverse corpora of texts, and do not appear to be an artifact of secular time trends.

\section{Methodology} 

\subsection{\textbf{Data}} 
Our research investigates three {\revision genres} of political speech: presidential debates in general elections since 1960, State of the Union (SOTU) speeches since 1961, and a sample of campaign speeches assembled by the American Presidency Project~\citep{woolley1999american}. Debates and SOTU speeches are generally standard across presidents and time, so all available documents are included in our main analyses.\footnote{Additional data are used for model training. See \secref{apps:lm_training} for details.}
Publicly available campaign documents, however, are imbalanced and not comprehensive over the same time frame, so these corpora are limited to speeches delivered within one month of Election Day {\revisionTwo in every presidential election} since 2008.
\tbref{tab:data_stats} provides {\revision summary} statistics of the final datasets for which we present results.\footnote{{\revisionThree Within these data, Donald Trump spoke in five debates (3,610 sentences), four SOTU addresses (1,471 sentences), and 28 campaign speeches (7,488 sentences) between the years of 2016-2020. 
}}

\begin{figure*}%
\centering
    \includegraphics[width=\linewidth]{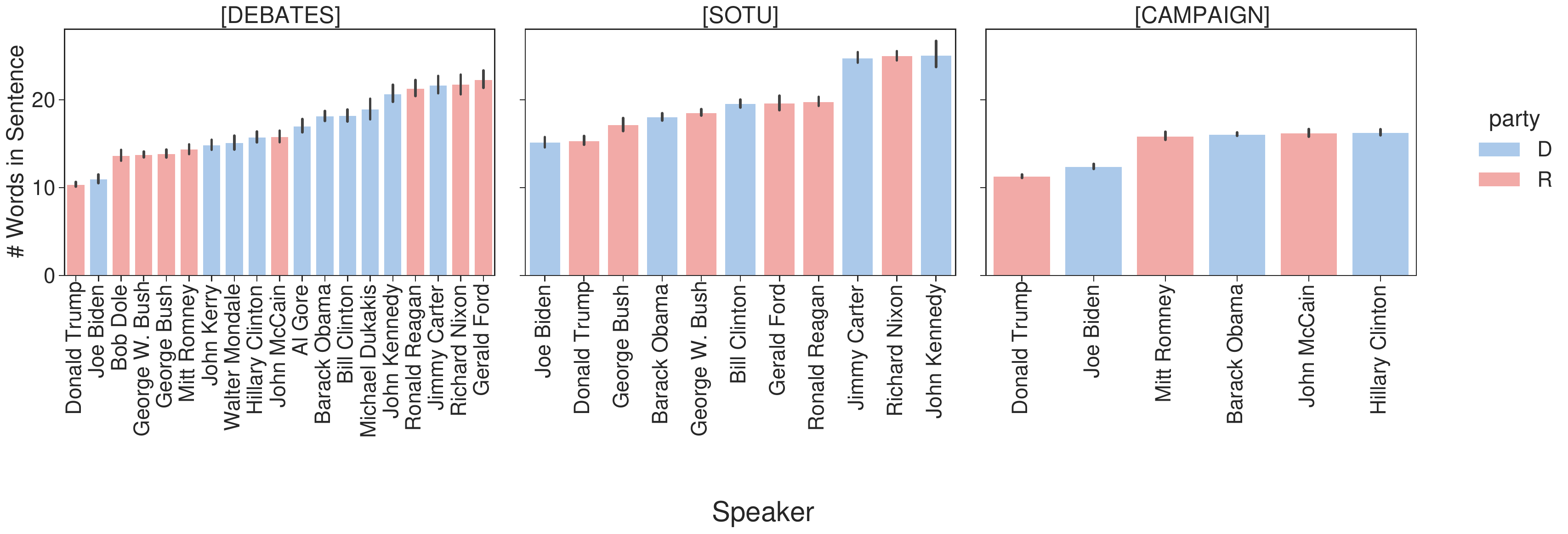}
\caption{Distribution of sentence lengths among all speakers, across debates, SOTU, and campaign data (error bars represent 95\% CI). Donald Trump tends to use the shortest sentences on average. 
}
\label{fig:sent_lens}
\end{figure*}

\figref{fig:sent_lens} shows the distribution of sentence lengths from each speaker across the datasets. Further details of how we collected these datasets can be found in \secref{apps:data} From the outset, however, we note that Trump tends to speak in markedly shorter sentences than do other presidents. Whereas Trump's sentences range from 10.4 to 14.5 words in the three data sources, the overall averages {\revisionTwo range} from 17.6 to 24.4 words. And when comparing presidents within each data source, Trump registered the single lowest number of average words per sentence among all presidents within debates and campaigns, and the second lowest number {\revisionTwo within} SOTU addresses.

\begin{table}[h]
\centering
\caption{Overall statistics of the three data types we use to present our main findings.
    }
    \label{tab:data_stats}
    
    \begin{tabular}{l r r r}
    \toprule
         &   DEBATES & SOTU  & CAMPAIGN \\
         \cmidrule(lr){1-4}
          
        \# speeches & 35  & 67  & 187 \\
        \# sentences & 35,096 & 22,775 & 36,295  \\
         date range  & 1960-2020 & 1961-2022  & 2008-2020 \\
         \# candidates compared & 19 & 11  & 6 \\
        party ratio (Dem:Rep) & 10:9 & 5:6 & 3:3 \\
          avg. sents/candidate &  1,449 & 2,070  & 5,701  \\
         avg. candidate sent len.   & 18.76 & 24.42   & 17.62  \\
         avg. speeches/candidate &  3.6 & 6.1  & 31.2  \\
    \bottomrule
    \end{tabular}
    
\end{table}

\subsection{\textbf{Quantifying the uniqueness of political speech}}

We quantify the uniqueness of political speeches using three complementary approaches: (i) a novel metric based on LLMs; (ii) a new resource of lexicons for divisive speech; and (iii) a comparative framework for {\revisionTwo the} portrayal of political opponents involving lexical features. Together, these three approaches allow for a robust and multifaceted comparison of speech patterns {\revisionTwo that involve global assessments beyond lexical comparisons and specific evaluations of divisiveness}. We briefly introduce the intuition behind each approach below.

 \paragraph{LLM-based uniqueness.} 
Large language models, like the GPT family of models, have received widespread attention for their abilities to statistically characterize the complex structures of natural language text. LLMs can measure the predictability of text by calculating the likelihood of the next word or sequence of words in a given context, to produce a measure known as {\revision ``perplexity,'' which is typically used to evaluate the quality of LLMs~\citep{melis2017state,chen1998evaluation}. To control for the length of a text, standard perplexity measures can be supplemented with measures of ``bits-per-character'' (BPC) \cite{liang2022holistic}.}

Contributing to this literature, we propose a metric of ``uniqueness'' based on the ability of LLMs to estimate the probabilities of word sequences, and we then use these estimates to compare political speech from various {\revisionTwo presidents and presidential candidates, whether delivered from the White House or on the campaign trail.} Specifically, from a pool of {\revisionTwo presidential candidates}, we determine how likely the speech of one speaker is to be produced by the others.  A positive value for one {\revisionTwo candidate's} sentence uniqueness, therefore, suggests that other {\revisionTwo candidates} are unlikely to say it. {\revision The larger this value, the less likely other candidates are to do so.}

The advantage of this metric is its consideration of the preceding context of given speech. {\revision Rather than examine words or phrases in isolation, this metric considers the order in which they appear, and thereby provides a much more nuanced characterization of speech patterns.} The precise meaning of the scores, however, {\revision may be difficult to interpret because they do not reveal the exact features that make a speech distinctive, and because they only allow for comparisons within a specified pool of speakers.}
Technical details on the construction of this metric are included in the \secref{apps:lm_training} and \secref{apps:uniq_score}.

\paragraph{Divisive speech lexicon.}
To analyze the actual content of language used by {\revisionTwo presidents and presidential candidates}, a ``divisiveness'' lexicon is created and applied to each dataset. We define language as ``divisive'' if it intends to impugn and delegitimize the speaker's target, e.g., by attacking their intelligence, integrity, or intentions. Examples of divisive accusations include ``racist'', ``dishonest'', ``corrupt'', or ``ridiculous''. 
Such labels are expressly designed to put the target on defense and accentuate differences and distance between parties.

Our definition of divisive is distinct from other commonly analyzed constructs such as political ``polarization'', which encompasses language that is associated more with one side than the other but is otherwise agnostic about its valence~\cite{10.1093/pnasnexus/pgac019}. Meanwhile, speech may be divisive without being ``toxic,'' which contains hateful, abusive, or offensive content \cite{Sap2021AnnotatorsWA}. Personal attacks can also be categorized as a form of toxic language, which prior work has examined primarily in the context of online, written communication \cite{wulczyn2017ex}.
\begin{revision}
    Nor is our measure of divisiveness the simple antonym of traditional notions of politeness~\cite{brown1987politeness}, which usually involve honoring social conventions, showing gratitude, paying compliments, avoiding complaints and curses, and respecting the listener's autonomy with the use of softening statements and hedges. More than just impolite or insulting, divisive speech, as we conceive it, is explicitly intended to serve the political purposes of delegitimization, marginalization, and distancing between speaker and target.
\end{revision}

{\revision To the best of our knowledge, ours is the first divisive speech lexicon. As we explain in further detail in \secref{apps:div_lex}, this lexicon consists of 178 words that four researchers independently reviewed to be qualitatively ``divisive'' in political speech. A strength of this lexicon-based analysis is its easy interpretability and applicability. Divisive words may be used by candidates from any political party in a wide variety of settings, are readily identified, and, as we subsequently show, are broadly agreed upon by coders. As with all lexical approaches, however, the measure is inherently limited by the subjective nature of lexical evaluations and the lack of contextual consideration.} 
\begin{revision}
    Our resource, like all lexicons, by itself does not account for surrounding context like negation or valence. Strategies like pairing our lexicon with part-of-speech taggers may be leveraged for downstream tasks.
\end{revision}

\paragraph{References to political opponents.} 
Our analysis further expands upon prior work by culling the subset of sentences that explicitly refer to political opponents. We specifically examine presidential debates, in which we define ``opponents'' here to be either the debate partner or their party.
The methodology of tagging opponent mentions is described in \secref{apps:tagging}.

Once speech referring to opponents is distinguished, we employ the Fightin' Words (FW) method \cite{monroe_fightin_2009} to identify words more strongly associated with opponent mentions; we extend this comparison across multiple candidates by calculating the overlap of each entity's corresponding word sets with those of all {\revisionTwo presidents and presidential candidates}.
Intuitively, a greater overlap indicates that other {\revisionTwo candidates} use similar rhetoric in opponent mentions, while a smaller overlap indicates that other {\revisionTwo candidates} do not use similar rhetoric in opponent mentions. 
This metric thus provides a novel measure for quantifying a {\revisionTwo candidate}'s distinctiveness with respect to their portrayal of opponents by combining lexical and graph analysis.
The results are readily interpretable and clarify the distinctive qualities of language used to characterize one's opponent. The metric is similarly constrained by the pool of {\revisionTwo candidates} and the finite selection of descriptors.

\section{Results}

\subsection{\textbf{Donald Trump is unique among all presidential candidates in all types of speech}}

\begin{figure*}[t]
\centering      
    \includegraphics[width=.85\linewidth]{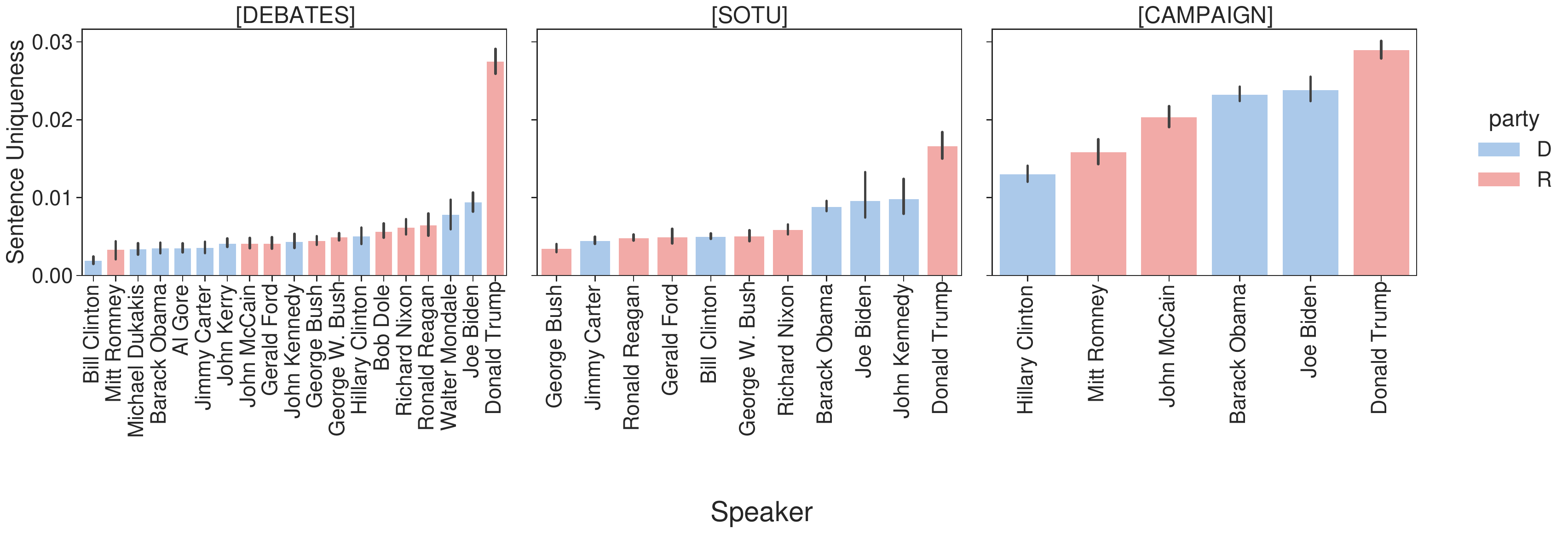}
\caption{ \label{fig:agg_uniq_overall}Average sentence uniqueness for each speaker, across all data types. Higher bars indicate greater uniqueness, i.e., that speaker's speech is less likely to be uttered by other candidates. Trump is the most distinctive speaker among these candidates for debates and SOTU speeches. Error bars correspond to the 95\% CI.}
\end{figure*}

\begin{figure}[t]%
\centering
\includegraphics[width=.95\linewidth]{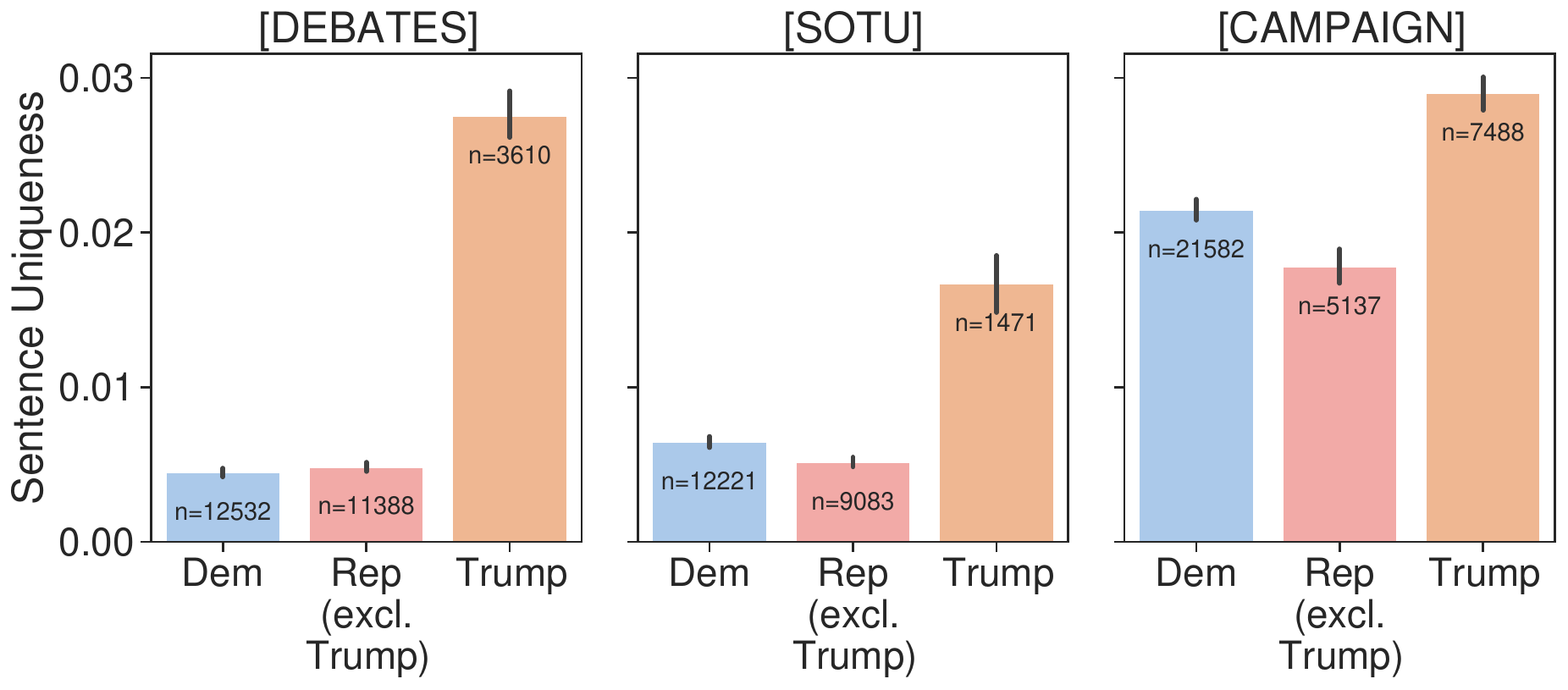}
\caption{Trump's average sentence uniqueness scores compared to the others aggregated by party. The number of sentences is denoted on each bar (with error bars representing 95\% CI).
Trump is significantly distinct from other Republicans. The uniqueness scores of the remaining Republicans are closer to those of Democrats than with Trump. For SOTU and campaign speeches, Republicans (without Trump) are less unique than Democrats on average. 
}
\label{fig:agg_uniq_party}
\end{figure}

\paragraph{LLM-based uniqueness.}
We start by presenting results based on the {LLM-based uniqueness} metric.
We find that Trump is the most distinctive speaker in debates and SOTU speeches (see Figs.~\ref{fig:agg_uniq_overall} and \ref{fig:agg_uniq_party}). Additionally, among campaign speeches analyzed from 2008 onwards, Trump speaks in ways that stand apart from all other candidates. 
By comparing candidate speech patterns aggregated by party, \figref{fig:agg_uniq_party} shows that Trump is more distinctive than his fellow Republicans for all types of speech. Indeed, the observed difference between Democratic and Republican candidates is minor compared to the gap observed between Trump and everyone else.

\begin{figure*}[t]
\centering
    \includegraphics[width=.85\linewidth]{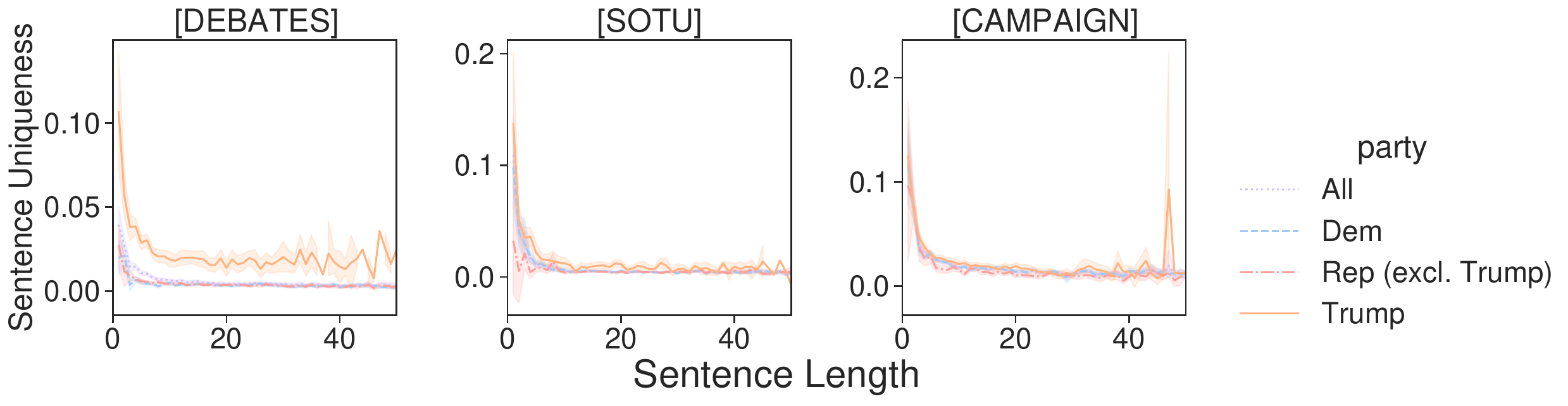}
\caption{\label{fig:agg_uniq_len_0_50} Sentence uniqueness across different sentence lengths, for Trump, all other Republicans, and Democrats. For debates and SOTU, Trump is consistently more distinctive across all sentence lengths.}
\end{figure*}

While Trump's speech is characterized by shorter sentences (\figref{fig:sent_lens}), we confirm that Trump's uniqueness is consistent across sentences of all lengths in \figref{fig:agg_uniq_len_0_50}. We also note that Biden has similarly short sentences on average as Trump, but his uniqueness scores are close in magnitude to those of the other candidates, particularly in debates and SOTU {\revisionTwo addresses}.

Furthermore, we find that there is minimal correlation between our uniqueness metric and standard simplicity scores, suggesting that the language model is not conflating uniqueness with language complexity (see 
\secref{apps:readability}
).

\paragraph{Further extensions and robustness checks.} When aggregating over presidential terms and years, Trump appears as slightly more distinctive in 2016 than in 2020 for debates, while his SOTU and campaign speeches increase in uniqueness over the years. Overall, he remains consistently more distinctive in both election cycles than the other candidates under consideration (see  \secref{apps:time_trends}). 
Prior to Trump's first election cycle, there are no clear temporal trends to suggest that uniqueness has been increasing over time.
Moreover, when comparing the uniqueness scores of the top decile of unique sentences for each speaker, we again find that Trump is the most distinctive speaker in all our samples of political speech (see \secref{apps:decile} ).

\begin{figure*}[t]
\centering
    \includegraphics[width=.85\linewidth]{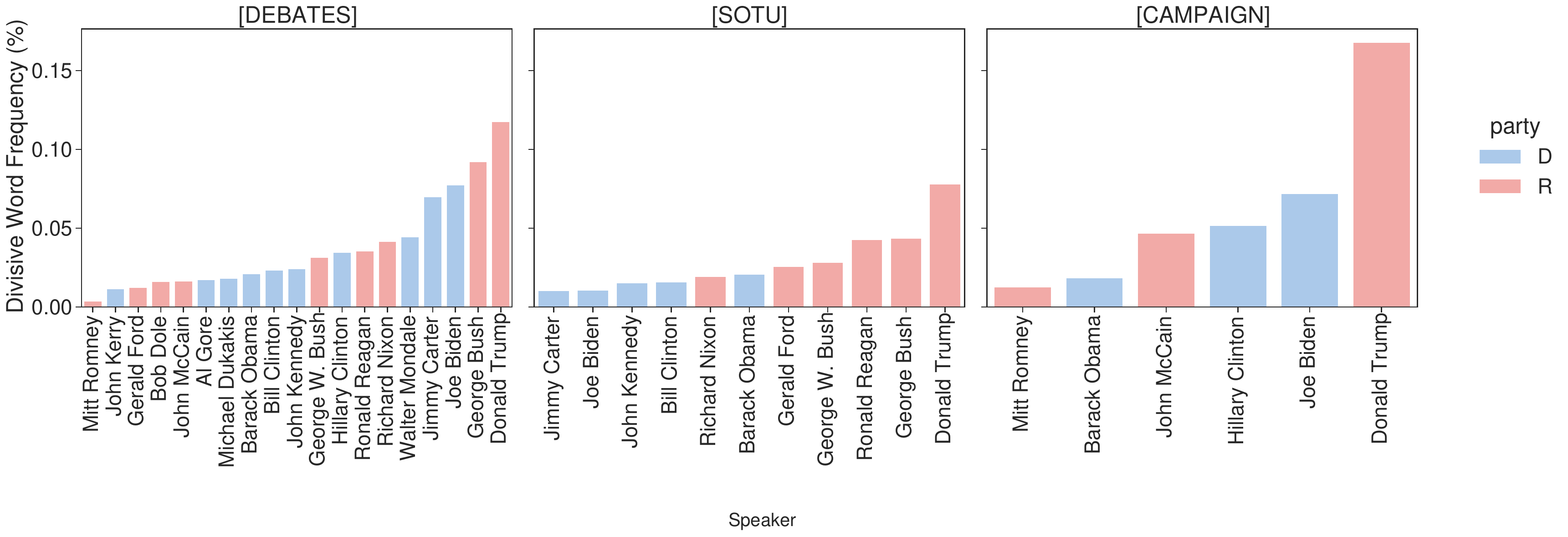}
\caption{\label{fig:main_figure_lexicon_a} Overall percentage of words used that are divisive. Trump uses the most words from our divisive lexicon, in all types of speech.}
\end{figure*}

\begin{revision}
\paragraph{Score validation.} The results presented in the main paper are based on the OpenAI GPT-2 model, which was the state-of-the-art open-source causal language model at the start of this project \cite{radford2019language}. We are interested in causal language modeling, i.e., the predictability of text in a \textit{forward sequential manner}. 
Causal language models are more appropriate for our task of scoring uniqueness, as opposed to masked language models (e.g., BERT, RoBERTa) which are designed to predict missing tokens within a sequence based on \textit{bidirectional} context (and thus excel at text classification).  
We validate our results with more recent and powerful LLMs like Gemma 2B \cite{team2024gemma} and Phi1-5b \cite{textbooks2} (see \secref{apps:valid_llm}). The results from these recent LLMs reaffirm our original findings; that is, Trump is consistently identified as the most unique speaker among modern presidential candidates in all types of speech.

\end{revision}

\subsection{\textbf{Trump speaks most divisively}}

\paragraph{Divisive word lexicon.} 
Leveraging our divisive word lexicon, we measure its usage across the different types of presidential speech. The frequency of lexicon word appearances is calculated for each candidate across each dataset and is shown in \figref{fig:main_figure_lexicon_a}. Word frequencies are calculated as the number of times a divisive word from the lexicon is spoken by a speaker in a given dataset (debates, SOTU, campaign speeches) divided by the total number of words spoken by the speaker in that in dataset. Both the lexicon words and the speech data are processed to remove contractions and punctuation.
Frequency trends provide a macroscopic look at the usage of divisive language over time. In \secref{app:div_results}, 
temporal plots and heatmaps of divisive word usage provide a more granular look and do not show strong trends over time.

\begin{table}[t]%
    \centering
\caption{Top 10 divisive words used by Trump in each type of speech. Raw counts of usage are denoted in parentheses next to each word.}

    \begin{tabular}{p{2.4cm} p{2.4cm} p{2.4cm}}
    \toprule
            DEBATES & SOTU  & {CAMPAIGN} \\
         \cmidrule(lr){1-1}\cmidrule(lr){2-2}\cmidrule(lr){3-3}
           stupid (14), \newline racist (14), \newline disgrace (12), \newline corrupt (8), \newline disgraceful (8), \newline ridiculous (6), \newline ashamed (6), \newline stupid (6), \newline filthy (6), \newline dishonest (6)
          &  cruel (3), \newline vile (3), \newline ruthless (2), \newline foolish (2), \newline corrupt (2), \newline reckless (2), \newline savages (1), \newline ugly (1), \newline outrageous (1), \newline ridiculous (1)
          & crazy (135), \newline corrupt (111), \newline stupid (69), \newline dishonest (53), \newline disgrace (45), \newline ridiculous (31), \newline racist (27), \newline incompetent (25), \newline stupidity (22), \newline ashamed (22) \\
    \bottomrule
    \end{tabular}
    
    \label{tab:trump_div}
\end{table}

One common trend shown by the frequency plots for each dataset is Donald Trump's relatively high usage of words from the divisiveness lexicon compared to other speakers. In all three datasets, Donald Trump's speeches rank highest in divisive word usage compared to the other candidates analyzed. Examples of Trump's most frequently used words from this lexicon usage are shown in \tbref{tab:trump_div}. He is most verbosely divisive in debates and campaign speeches, uttering terms like ``crazy'', ``corrupt'', and ``stupid'' with high frequency. 
The contexts of debates and campaigns are more combative than SOTU addresses, which is reflected in the higher levels of divisive language in those two mediums. \tbref{tab:trump_div_sents} contains a selection of Trump's divisiveness in context; additional example sentences from various speakers can be found in the \secref{app:div_results}.

\begin{revisionTwo}
Across all debates and campaign speeches, which are the two corpora with greatest prevalence of divisive language, sentences that use divisive words are more unique (see \secref{app:comp_metrics}). The Spearman correlation coefficient between uniqueness and divisive word usage is 0.01 in debates and campaigns ($p<0.05$ in campaigns and $p=0.30$ in debates). 
The weak, positive correlation suggests that divisiveness amounts to only a small part of the overall uniqueness of any speech pattern.

\end{revisionTwo}

\begin{table}[t]%
    \centering
    \caption{Example sentences from debates and campaigns, with preceding context, spoken by Trump that use words from the divisive lexicon. Bolded words are
    matched with our divisiveness lexicon. %
    }\label{tab:trump_div_sents}
    \begin{tabular}{p{8cm}}
    \toprule
            DEBATES: \newline
           \textit{Trump:} Let me tell you something. \newline
\textit{Trump:} You take a look at Mosul. \newline
\textit{Trump:} The biggest problem I have with the \textbf{stupidity} of our foreign policy, we have Mosul.  \newline 
\textemdash \newline
\textit{Trump:} I was at a little Haiti the other day in Florida. \newline
\textit{Trump:}  And I want to tell you, they hate the Clintons, because what's happened in Haiti with the Clinton Foundation is a \textbf{disgrace}.
\\
           \cmidrule(lr){1-1}
           CAMPAIGN: \newline
\textit{Trump:} We're going to bring back the miners and the factory workers and the steel workers.\newline
\textit{Trump:} We're going to put them back to work.\newline
\textit{Trump:} The economic policies of Bill and Hillary Clinton have destroyed manufacturing in your state and throughout the entire country.\newline
\textit{Trump:} The \textbf{corrupt} Clintons gave us NAFTA. \newline
\textemdash \newline
\textit{Trump:} The fact is, this is the single most important election in the history of our country. \newline
\textit{Trump:} And sleepy Joe Biden's made a corrupt bargain. \newline
\textit{Trump:} You saw the bargain he made, in exchange for his party's nomination, which he shouldn't have gotten because if Pocahontas got out one day early, I'd be running against \textbf{Crazy} Bernie, which would have been okay, too.
\\
    \bottomrule
    \end{tabular}
    
\end{table}

\subsection{\textbf{References to political opponents in speech}}

In addition to examining overall speech patterns, we can also extract the portions that reference political opponents. Next, we show that Trump is more likely to mention his political opponents than are other presidential candidates and that when doing so, Trump uses particularly distinctive language.  

\paragraph{Rates of opponent mentions.} As one might expect, candidates routinely mention their opponents in debates and only rarely in SOTU addresses. Overall,  rates of sentences that mention opponents for debates, SOTU, and campaign speeches are 20.60\%,  0.83\%, and 6.95\%, respectively (see \secref{app:opp_results}). 
Trump has the highest rate of opponent mentions in debates. 

\begin{figure}[h]%
\centering
    \includegraphics[width=.95\linewidth]{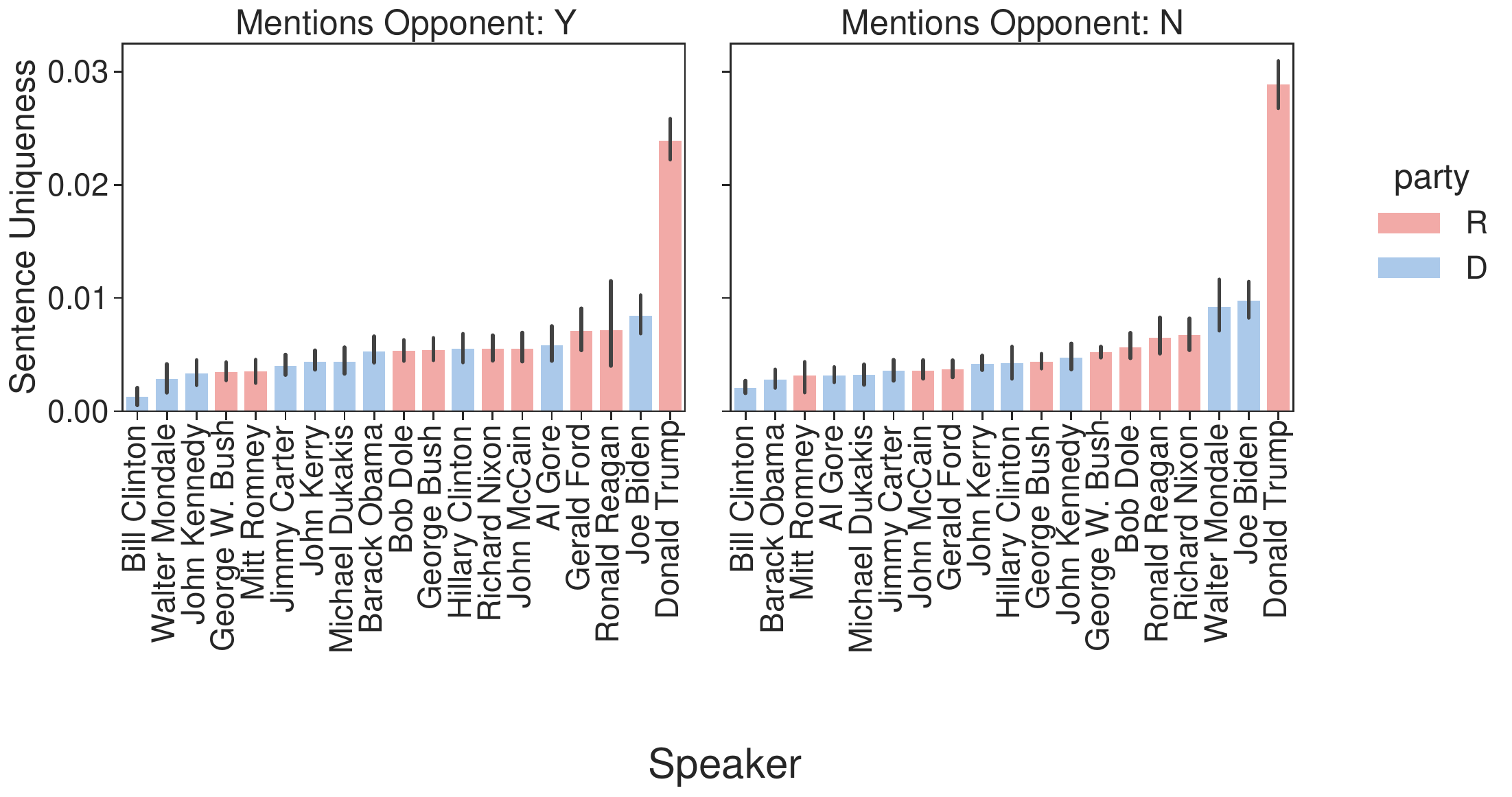}

\caption{Uniqueness of speech broken down by opponent mentions, for debates (the error bars represent 95\% CI). Trump is most distinctive regardless of whether he references an opponent or not.}
\label{fig:agg_uniq_opp_ments}
\end{figure}

Since the opponent mentions are more common in presidential debates, we revisit our LLM-based measure for the debates to confirm that Trump is not unique simply because he is more likely to mention opponents. \figref{fig:agg_uniq_opp_ments} shows that the LLM-based uniqueness score rankings are relatively consistent across opponent and nonopponent mentions. Moreover, Trump's language is consistently unique in these debates, whether or not he calls out an opponent. 
{\revisionTwo 
Concurrently, sentences that refer to an opponent tend to be more distinct than those that do not (see \secref{app:comp_metrics}). And in debates, we additionally find that sentences that mention opponents contain significantly higher frequencies of divisive words.  If you are looking to isolate the distinguishing characteristics of a candidate's speech patterns, therefore, you would do well to focus on how they talk about their political opponents.

}

\paragraph{Fightin' Words adjective overlap.}  
To identify which \textit{adjectives} are most strongly associated with describing opponents for each candidate, we calculate the odds ratios between opponent mention sentences and nonopponent mention sentences, i.e., Fightin' Words (FW) \cite{monroe_fightin_2009}.
After these odds ratios are calculated, we compare the words most commonly associated with opponent mentions across candidates and calculate an overlap score. This \textit{FW overlap metric} for each candidate is calculated as the average number of speakers who share each of the candidate's top-$n$ FW adjectives. 
See \secref{apps:fw_overlap} for further details on the construction of this metric.

Intuitively, the lower the FW overlap score for a candidate, the fewer adjectives that candidate uses in common with others when describing their political opponents. For example, a low FW overlap metric indicates that the words a politician uses most frequently to describe an opponent are distinct from those of other candidates.
From \figref{fig:fw_a}, we see that Trump has the lowest FW overlap in debates on average across different top-$n$ thresholds. Among adjectives that Trump uses in references to opponents, his top-$n$ FW are consistently distinct for $n=5$ to $n=25$.\footnote{There are considerably more sentences of nonopponent mentions, but only looking at top-$n$ FW (across different thresholds) helps control for that difference.}
\figref{fig:fw_b} ranks the speakers by their FW overlap score at the top-$15$ threshold and shows that Trump has the lowest FW overlap scores when mentioning opponents (additional top-$n$ plots can be found in the \secref{app:fw_overlap}). 
Trump has a higher overlap of FW descriptors when not referencing opponents, indicating that he is more distinctive for how he {\emph{describes}} his opponents. 

Examples of Trump's top-$25$ FW adjectives can be seen in \tbref{tab:trump_fw}; the left panel includes language used to describe his political opposition, while the right panel includes references to his own party and allies. In addition to 
attacking opponents (e.g., ``disgraceful''), Trump tends to use fairly simplistic adjectives like ``massive'' and ``super''.\footnote{Some of these words, like 
``radical'' and ``liberal,'' appear in the ``polarization'' dictionary of \cite{10.1093/pnasnexus/pgac019}. Taken as a whole though, our divisive word lexicon contains very little overlap with that dictionary.}

\begin{figure}
    \centering
    \includegraphics[width=\linewidth]{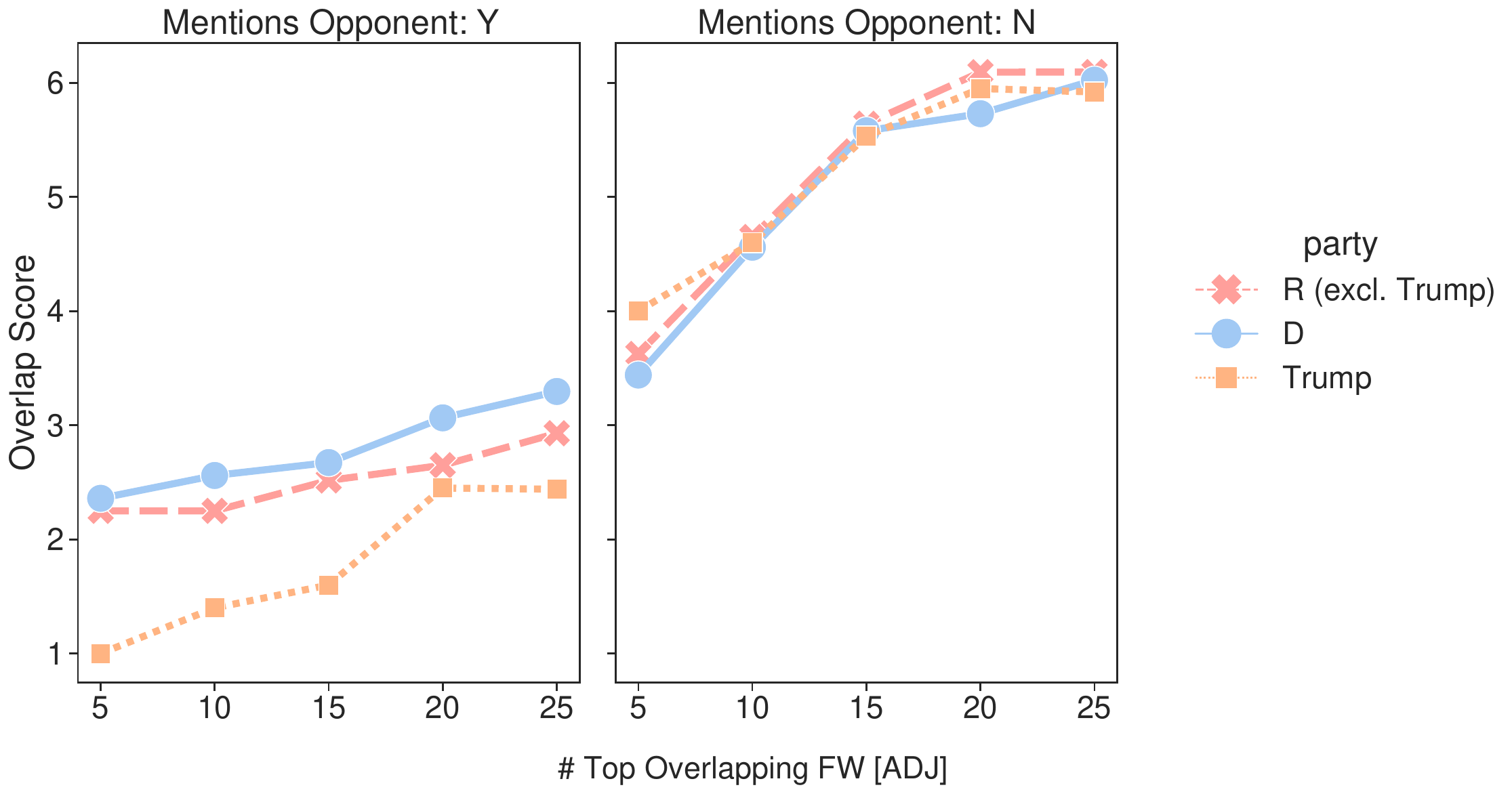}
\caption{\label{fig:fw_a} FW overlap scores across different top-$n$ thresholds, comparing Trump and the other candidates aggregated by party.}
\end{figure}

\begin{figure}
    \centering
    \includegraphics[width=\linewidth]{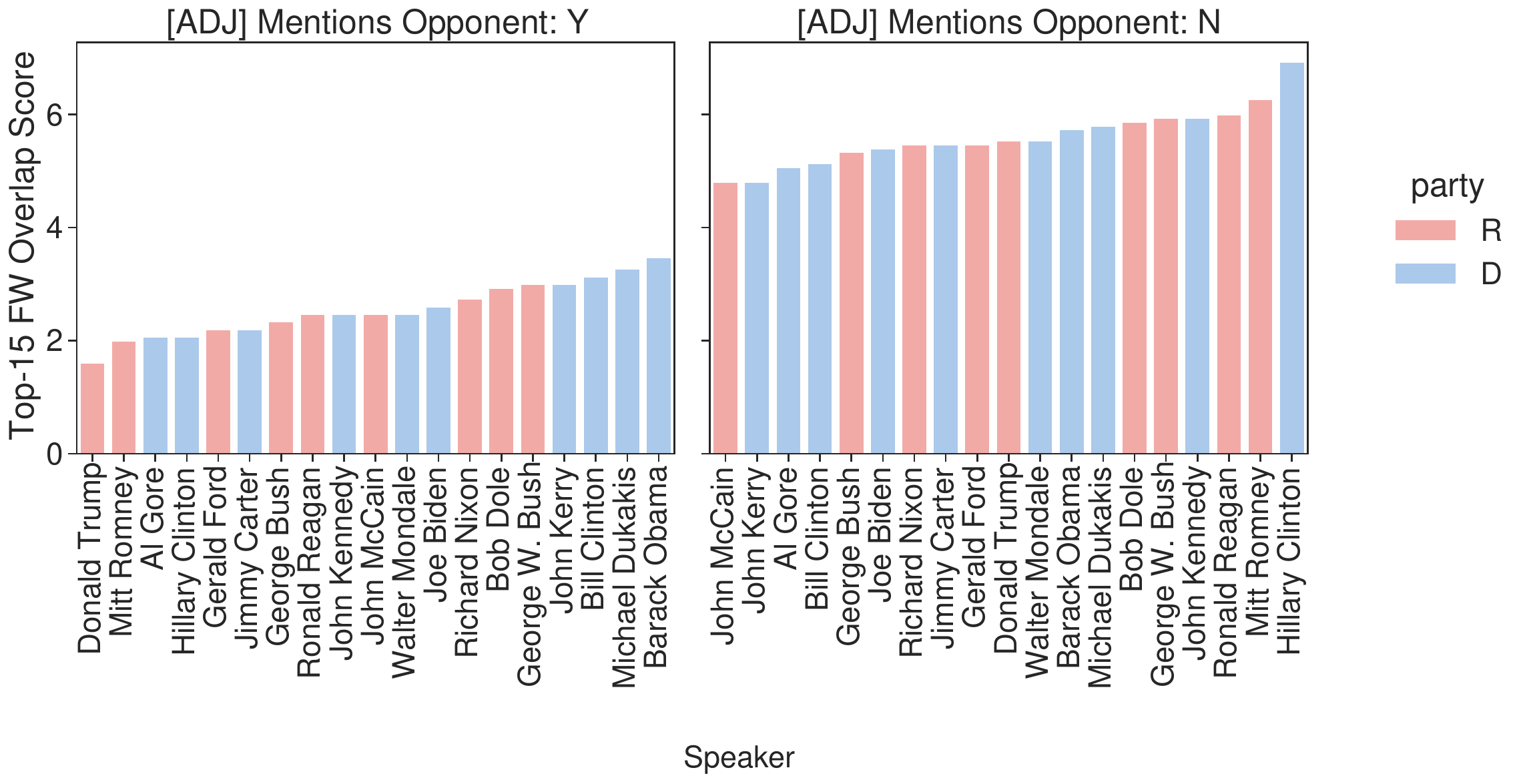}
    \caption{\label{fig:fw_b} Top-15 FW overlap score in debates for each candidate. Trump's FW associated with opponent mentions generally have the lowest overlap in adjective usage compared to other candidates, which is another indication that his language is distinctive. }
\end{figure}

\begin{table}[h]%
    \centering
    \caption{Top-25 FW adjectives said by Donald Trump in debates, either in reference to opponents or not. Z-scores are provided in parentheses. 
    }
    \label{tab:trump_fw}
    \begin{tabular}{p{3.75cm} p{4cm} }
    \toprule
            Mentions Opponents: Y & Mentions Opponents: N   \\
    \cmidrule(lr){1-1} \cmidrule(lr){2-2}
           left (2.07), long (1.85), tough (1.80), own (1.79), ok (1.79), bigger (1.79), worse (1.66), radical (1.57), super (1.50), real (1.50), effective (1.47), xenophobic (1.46), liberal (1.46), disgraceful (1.46), massive (1.33), single (1.33), political (1.09), last (1.09), economic (1.08), short (1.08), huge (1.04), various (1.04), red (1.04), upset (1.04), back (1.04)
          &  good (-3.63), great (-3.25), important (-2.36), more (-2.08), inner (-2.00), strong (-1.99), right (-1.84), proud (-1.80), other (-1.79), expensive (-1.64), old (-1.53), big (-1.53), better (-1.50), greatest (-1.44), young (-1.41), sad (-1.41), able (-1.31), much (-1.31), african (-1.27), fine (-1.27), beautiful (-1.27), tougher (-1.18), nice (-1.18), least (-1.12), sure (-1.12)
          \\
    \bottomrule
    \end{tabular}
    
\end{table}

\section{Discussion}

{\revisionThree In this work, we proposed novel approaches to quantify the uniqueness and divisiveness of presidential discourse.}
{\revisionThree Our results show that whether on the campaign trail, the debate stage, or the official dais of the House of Representatives, Trump speaks differently from all modern {\revisionTwo presidents and} presidential candidates.} We confirm his distinctiveness for speaking in shorter, simpler, and more repetitive sentences. We further quantify the uniqueness of his overall speech compared to that of other presidential candidates. Finally, we demonstrate that Trump uses language that is more divisive, antagonistic, and explicitly focused on his political opponents. 

The differences between Trump and other candidates do not appear to be an artifact of secular communication trends (see \secref{apps:time_trends}),
whether in the general coarsening of public discourse or the tendency to speak in simpler sentences. Across multiple points of comparison, observed differences between Trump and other contemporary candidates appear every bit as large as those between Trump and candidates from the 1960s and 1970s.

Our findings, of course, come with a variety of limitations. The campaign data, for instance, include samples from only the most recent candidates and exclude certain communication formats, such as interviews. None of our datasets cover instances when surrogates (family members, vice presidents, etc.) speak on behalf of a candidate. 
\begin{revision}While other works have examined Trump's tweets \cite{Lewandowsky_Jetter_Ecker_2020}, we do not assess any of his social media postings for lack of comparable data for other candidates, especially older ones. \end{revision} Furthermore, our data collection ends at 2022, excluding more recent remarks from Trump that have been flagged in the media for their escalated divisiveness \cite{Kurtzleben_2023}.

{\revisionTwo Our research invites a variety of extensions.
Given that values of the LLM-based uniqueness metric can only be understood relative to the elements of a selected sample, future work would do well to compare Trump to other speakers, such as populist leaders in other countries.
Just as we investigate divisiveness as a new dimension of political speech with particular resonance for Trump's rise to power, other scholars might evaluate other dimensions of political speech, such as its analytical sophistication or propensity to endorse racial, gender, or sexual stereotypes. Additionally, future research might explore the possibility of contagion effects from Trump's speech patterns to other political figures, as politicians periodically emulate and influence one another.
} 

We also recognize that all of our metrics come with tradeoffs{\revisionTwo, which we have elucidated throughout this essay.} Still, the tools we develop can be applied to a wide variety of settings. While we use them to compare speech patterns among presidents and presidential candidates, they can just as easily be deployed to analyze the language used by any public officials. And in addition to characterizing overall differences in speech patterns, we illuminate ways of assessing particular qualities that appear in either stand-alone speeches or interactive exchanges.

The substantive findings presented herein, moreover, establish the {\revisionTwo uniqueness} of Trump's speech patterns, just as they reveal particular qualities that distinguish Trump from all modern US presidential candidates. More research{\revisionTwo, of course,} is required to map these speech patterns into larger political strategy. Nonetheless, we conjecture that these qualities broadly contribute to Trump's enduring appeal as a populist who unabashedly denounces established political enemies in a historical period of acute polarization, distrust, and division.

\section*{Acknowledgments}
This work is in part supported by a Data \& Democracy seed funding at the University of Chicago.

\section*{Data availability}
The data underlying this article are available in the American Presidency Project at \url{http://www. presidency. ucsb. edu/ws}, and can be accessed publicly.
\begin{revision}
    For reproducibility, we release our data and code at \url{https://github.com/ChicagoHAI/quantifying-unique-and-divisive-speech}.
\end{revision}

\bibliography{refs}%

\begin{appendices}

\section{Implementation Details}
\label{app:mats_methods}

\subsection{\textbf{Data collection}}\label{apps:data}

Each dataset is scraped from the American Presidents Project (APP) database \cite{woolley1999american}. In general, all text for each speech is scraped and tagged with its speaker. In addition, some metadata are collected, like the date and title of the speech or document. In order to properly annotate speakers, particularly in the debate speeches, the speaker's name is identified based on a variety of factors like special text styling, string matching, and html formatting. In addition, audience annotations, like ``laughter'' and ``applause'',  are filtered out from the data to the best of our ability. After scraping, the data are subsampled and reviewed to ensure quality and correct speaker annotations.

\paragraph{Debates data} 
The Presidential debates data include General Election debates from 1960 to the present day and Primary Election debates since 2000. For the purposes of this analysis, we exclude primary debates and any general election debates that featured vice-presidential candidates. Sentences spoken by candidates who were not the nominee for either the Republican or Democratic party are also excluded (e.g., Ross Perot in 1992, moderators, and audience members).

\paragraph{SOTU data}
Because we focus on modern presidents, State of the Union addresses from 1961 onwards are used in deriving our main results. The SOTU dataset includes a few speeches that are not officially considered State of the Union addresses but functionally operate as such. Beginning with Reagan, recent presidents have started addressing a joint session of Congress shortly after their inaugurations. According to the American Presidency Project, ``it is probably harmless to categorize these as State of the Union messages (as we do). The impact of such a speech on public, media, and congressional perceptions of presidential leadership and power should be the same as if the address was an official State of the Union''.\footnote{\url{https://www.presidency.ucsb.edu/documents/presidential-documents-archive-guidebook/annual-messages-congress-the-state-the-union}} 

\paragraph{Campaigns data}
All campaign documents from 1932 to 2020 are collected initially.
 We manually identify keywords that indicate campaign speeches of interest and filter speech titles based on those keywords. Keywords selected are: ``remarks'', ``speech'', and ``address''. 
Manually removed documents include: speeches with reporters, question/answer format speeches, press releases, town halls, press releases, co-appearances with other politicians and spouses, etc.

Subsequently, we apply clique filtering to ensure that any duplicate speeches are removed. It is common that a candidate has a stump speech that is delivered at multiple events, and to diversify and balance the dataset, we remove such duplicates. These de-duplicated data are used for model training.

After filtering, our final campaign dataset includes only speeches delivered within the month before election day and only from candidates since 2008. This smaller subset is used to generate the results presented in the main paper.

\subsection{\textbf{Sentence and opponent mention tagging}}
\label{apps:tagging}
After each dataset is filtered, sentences are tagged as mentioning an opponent or not through an automated process. Sentences are classified as definitely including an opponent mention, possibly including an opponent mention, or not including an opponent mention based on keywords and the presence of parts of speech. For debates, ``opponents'' are identified by the name(s) of the debate partner or their party. After automatically tagging the debates dataset, the sentences that are labeled as possibly including an opponent mention are manually reviewed by an expert team of four researchers. The team compares a subset of pairwise overlapping ratings for consistency. The Cohen's $\kappa$ for inter-coder agreement is roughly 0.8, which indicates substantial agreement.

For SOTU, opponent mentions only include those of other presidential candidates; names of candidates who never held presidential office are not categorized as mentions of opponents, in contrast to the debates dataset. Following a manual review of selected speeches from the SOTU dataset, it is concluded that references to nonpresidential figures are much more neutral and considerably less frequent in State of the Union speeches than they are in debate contexts.

For campaign data, we automatically tag opponents using the names of the final party candidates from the opposing party of the speaker. Opposing-party candidates from the primaries are not automatically tagged. 

The strict guidelines for automatic tagging ensure high precision of the labels. Limitations of this automatic tagging method include lack of coreference resolution and difficulty of measuring recall.

\subsection{\textbf{Language model training and analysis}}
\label{apps:lm_training}
We fine-tune a pretrained, {\revision 124M parameter} GPT-2 model using the huggingface \cite{radford2019language} and PyTorch Lightning \cite{Falcon_PyTorch_Lightning_2019} libraries for each of the three data types, resulting in three different language models:
\begin{itemize}
\item \LMdebates, trained on 35 debates (35,096 sentences) from 1960 to 2020%
\item \LMsotu, trained on 246 speeches (69,630 sentences) from 1790 to 2022 %
\item \LMcampaign, trained on 640 documents (83,038 sentences) from 1932 to 2020 %
\end{itemize}

To preprocess data for training, we parsed each speech into sentences, prefixed each sentence with the speaker prompt (e.g., ``Donald Trump:'', and masked any named entities (as identified by spaCy NER tagger) with a \texttt{<ENT>} mask.\footnote{Versions of each model were also trained on unmasked data. The results from these models are consistent with our main findings and are presented in \secref{apps:unmasked_model}.})
We fine-tune each model for 10 epochs on all available corresponding data with a learning rate of $5e$-$5$. {\revision Validation with larger, more recent LLMs is shared in \secref{apps:valid_llm}. }

\subsection{\textbf{BPC/predictability and uniqueness scores}} 
\label{apps:uniq_score}
Consider a set of presidents or candidates $C= \{c_1, \dots, c_n\}$ who each have a corresponding set of sentences $\mathcal{S} = \{S_1, \dots, S_n\}$ where each set $S_i = [s_i^1, \dots, s_i^m]$, for each of the three data types.

We use bits-per-character (BPC), also known as bits-per-byte, as a proxy for ``predictability'' of a sentence. Lower BPC values correspond to higher predictability. 
We use BPC instead of perplexity or loss directly to account for variation in tokenization techniques. 

 We calculate BPC as follows.
 From our fine-tuned models, we are able to obtain {\em loss} values, $\mathcal{L}(t_i)$ for each token $t_i$ in an input. 
For some sentence of tokens $s = (t_1, t_2, \dots, t_k)$ where $\text{len}(s)$ denotes the number of characters in $s$:
\[ \mathcal{BPC}(s) = \frac{1}{ \text{len}(s)} \sum_{i=1}^{k} \mathcal{L}(t_i)  \]

In other words, $\mathcal{BPC}(s)$ is the sum of cross-entropy losses of each token in a sentence $s$, divided by the number of characters (bytes) in the $s$.
{We calculate the $\mathcal{BPC}(s)$ of a sentence $s$ with a context window size of 512 tokens. That is, preceding sentences of the one in question are provided to obtain the most representative score of its predictability.}

We first obtain the BPC of a sentence $s_i^j$ with its original speaker $c_i$ as its speaker prompt. For example, ``Donald Trump: Make America great again'' is denoted by $\mathcal{BPC}_{c_i}(s_i^j)$. The BPC of the same sentence $s_i^j$ with an alternative speaker prompt $c_k$, e.g., ``Hillary Clinton: Make America great again'' is denoted by $\mathcal{BPC}_{c_k}(s_i^j)$.
Now, we define a ``uniqueness'' score for $s_i^j$ sentence as follows:
\[\textsc{SentUniq}(s_i^j) = \left( \frac{1}{|C|-1} \sum_{c_k \in C \setminus \{ c_i\}} \mathcal{BPC}_{c_k}(s_i^j) \right) - \mathcal{BPC}_{c_i}(s_i^j)\]

Intuitively, this means that the uniqueness of a sentence is defined as the difference between its BPC with its original speaker prompt and the average of its BPC scores with each of the $|C|-1$ alternative candidates as a speaker prompt. The greater this difference is, the higher the $\textsc{SentUniq}(s_i^j)$ score is, suggesting that sentence $s_i^j$ is most likely to be said by the original speaker $c_i$ and not by the other candidates. Since debates include an opponent speaker in the transcript (e.g., Trump vs Biden), we exclude the opponent from the replacement candidates for sentences from that particular debate (i.e., $ C \setminus \{\text{Trump, Biden}\})$, to avoid simulating unrealistic debates with only one speaker.

Then, the overall ``uniqueness'' score for a candidate $c_i$ is the average of the $\textsc{SentUniq}$ scores for each of $c_i$'s sentences in $S_i$:

\[\textsc{Uniq}(c_i) = \frac{1}{|S_i|} \sum_{s_i^j \in S_i} \textsc{SentUniq}(s_i^j)\]

Again, the intuition here is that the larger this score, the greater the average difference is for sentence uniqueness, i.e., overall, speaker $c_i$'s sentences are not likely to be said by another candidate.

\subsection{\textbf{Divisiveness lexicon}} 
\label{apps:div_lex}
First, a vector space word model is used to populate a list of candidate divisive words. Then, we review and refine the candidate list through researcher annotations. The word vector model used is Gensim’s glove-wiki-gigaword-300 \cite{rehurek_lrec, pennington-etal-2014-glove}.

Ten seed ``divisive'' terms are chosen manually by NLP and political science experts as common politically divisive English terms, before seeing the actual speech data. 
The ten seed terms used are: ``stupid'', ``dishonest'', ``unamerican'', ``idiot'', ``deplorable'', ``pathetic'', ``immoral'', ``disgrace'', ``incompetent'', ``foolish''. With these seed terms, 350 additional terms with the highest cosine similarity in the vector space model are added to the lexicon. Each of these initial 360 terms is analyzed by four researchers as ``divisive''or  ``not divisive'' based on the following criteria:
\begin{itemize}
  \item No modifiers, e.g., ``utterly'', ``extremely''
  \item Should not include words that can often be used both divisively and nondivisively
  \item  Should only include words that would be considered divisive in most political contexts
\end{itemize}
Only words that receive a majority of votes by the annotators are included in the final lexicon. 
The final size of the lexicon is 178 words. The entire lexicon can be found in the \secref{app:div_lexicon}. 
\begin{revision}
\paragraph{Annotator agreement.} 
Following \cite{mohammad-turney-2010-emotions}, in order to analyze how often the annotators agree with each other, we additionally calculate the percentage of times the majority class has size 4 (all annotators agree), size 3 (all but one agree), and size 2 (even split). \tbref{tab:annot_majority} shows these agreement values: 58.9\% of the 360 initial terms had full annotator agreement, and 86.7\% of these 360 have at least 3/4 annotators agreeing on the label. When limiting to the 178 terms that at least 3 out of 4 annotators agree are divisive, 69.1\% have full agreement on the label (and 100\% have majority agreement, by definition). So, the final terms include primarily instances that have high agreement scores.

\begin{table}[h]%
    \centering
\caption{{\revision Percentage of majority class annotator agreement on binary labels (divisive? ``yes'' or ``no''). 58.9\% of the 360 initial terms had full annotator agreement, and 86.7\% of these 360 have at least 3/4 annotators agreeing on the label. When limiting to the 178 terms that at least 3/4 annotators agree are divisive, 69.1\% have full agreement on the label.}}

    \begin{tabular}{p{2.5cm} c c c}
    \toprule
     & \multicolumn{3}{c}{Majority class size (\%)} \\
           Terms & four & three  & two\\
            \cmidrule(lr){1-4} 
            360 initial & 58.9 & 27.8 & 13.3\\
            178 final & 69.1 & 30.9 & 0.0 \\
    \bottomrule
    \end{tabular}
    
    \label{tab:annot_majority}
\end{table}

Since there are four annotators, we calculate Fleiss' $\kappa$ for interannotator agreement, as opposed to Cohen's $\kappa$ which is designed for two raters. For the initial 360 terms, we obtain a Fleiss' $\kappa$ of $0.54$, which indicates a moderate level of agreement. Discussion of this score and why it may underestimate true agreement is included in \secref{app:annot_agreement}.
\end{revision}

\begin{figure}[bthp]
\centering

    \includegraphics[width=.65\linewidth]{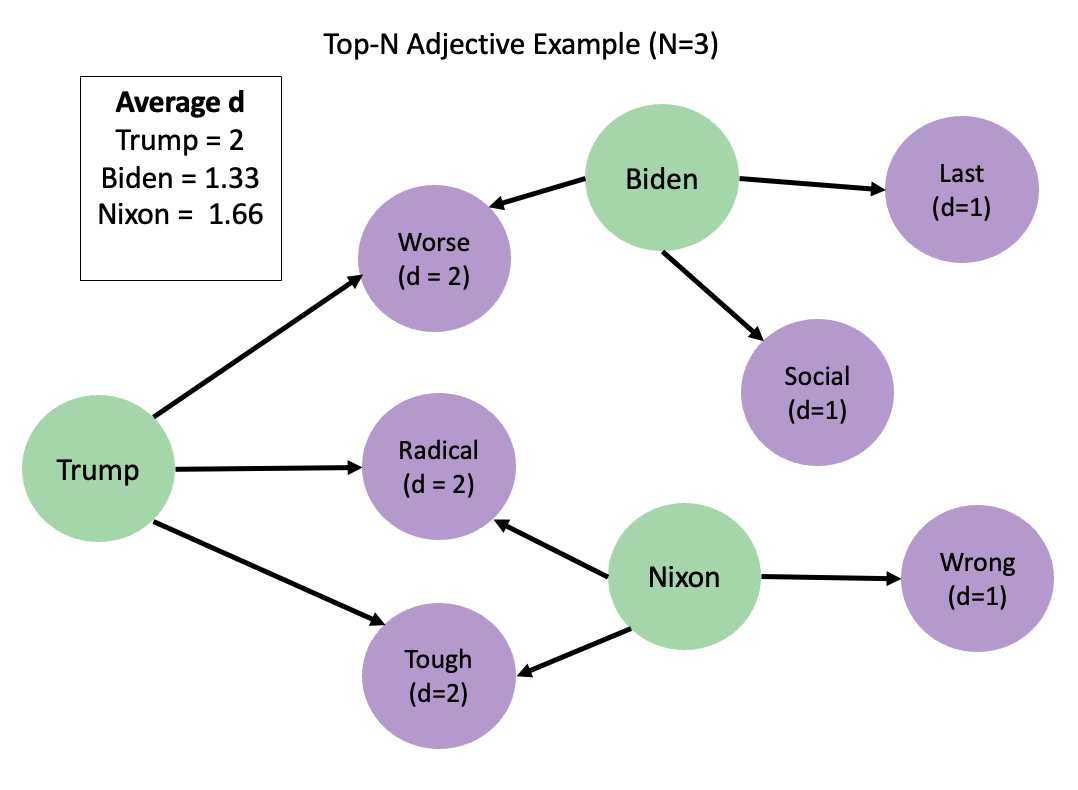}
\caption{Conceptual example of how the FW overlap metric is calculated for $N=3$.}
\label{fig:fw_overlap_ex}
\end{figure}

\subsection{\textbf{Fightin' Words overlap metric}}  
\label{apps:fw_overlap}
Monroe et al.~\cite{monroe_fightin_2009} introduce a methodology for lexical feature selection that calculates odds ratios between of word probabilities between two related corpora. We use this method with the informative Dirichlet prior to obtain the Fightin' Words (FW) in a data type for each candidate. 

For each candidate $s \in C$, we specifically get the set of FW between the words associated with their opponent mentions ({FW}$_Y(s)$) versus the set of those that are not ({FW}$_N(s)$). We then examine the top-$n$ of these sets respectively, {FW}$_Y^n(s)$ and {FW}$_N^n(s)$.
For $Y$ opponent mentions, we create a graph representation as follows (see \figref{fig:fw_overlap_ex}):
\begin{itemize}
    \item speaker nodes $S$, where each node corresponds to a candidate
    \item word nodes $W$, where $W$ corresponds to the union of each candidate's top-$n$ ({FW}$_Y^n(s)$), i.e., $W = \bigcup_{s \in C} {FW}_Y^n(s)$
    \item edges $E(S, W)$, where edge $e$ is added between $s\in S$ and $w\in W$ if $w$ is in {FW}$_Y^n(s)$
\end{itemize} 
The top-$n$ FW overlap metric ($OM_Y^n(s)$) is then calculated as the
\[ OM_Y^n(s) =  \frac{1}{n} \sum_{\forall w \in FW_Y^n(s)} \operatorname{deg}(w),\]
where $\operatorname{deg}(w)$ corresponds to the degree of the word node (\# edges entering $w$).
Likewise, for the set of words not associated with opponent mentions, we have
\[ OM_N^n(s) =  \frac{1}{n} \sum_{\forall w \in FW_N^n(s)} \operatorname{deg}(w).\]

Equivalently,  the overlap metric for $s$ can be thought of as the average number of speaker sets {FW}$_Y^n(c_i),~\forall c_i \in C$ that the top-$n$ words of $s$ appear in. A low $OM_Y^n(s)$ indicates that speaker $s$ uses more distinct language to refer to opponents, while a higher score corresponds to opponent-referring language that is similar to that of other candidates.

\section{LLM-based Uniqueness Robustness Checks}

\subsection{BPC} 
\begin{figure}%
\centering
\begin{subfigure}{.95\linewidth}
\centering
    \includegraphics[width=\linewidth]{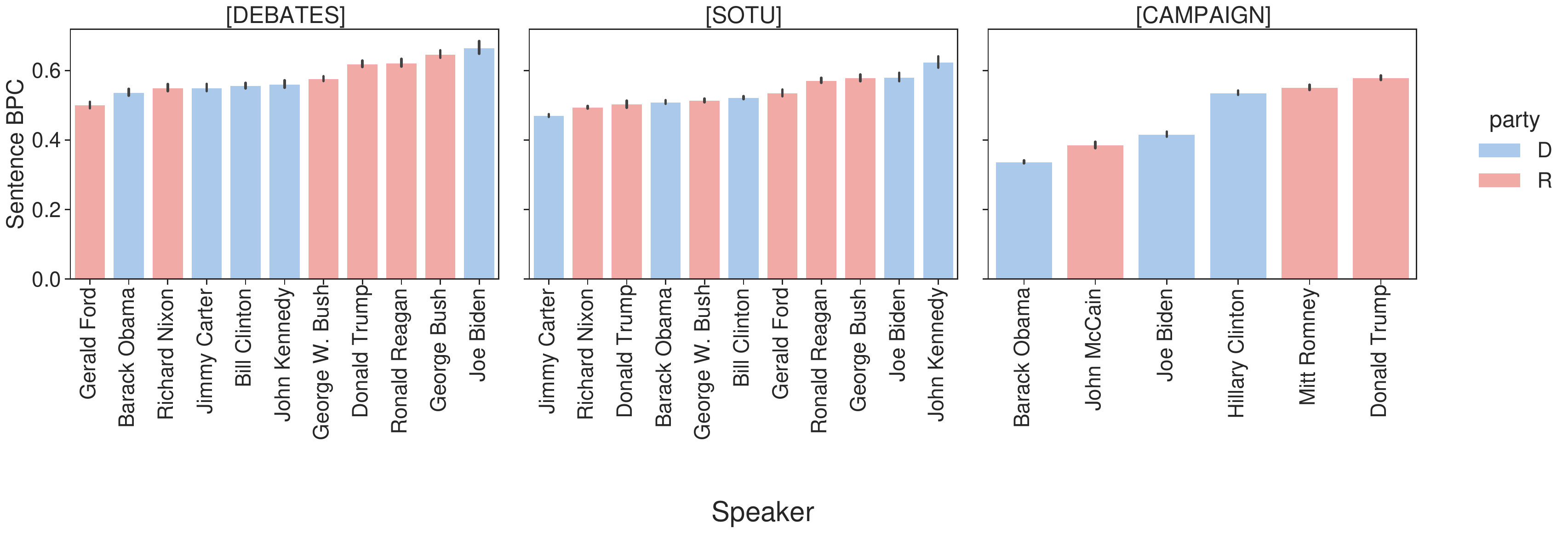}
    \caption{Average sentence BPC per candidate (error bars represent 95\%-confidence intervals)}
    \label{fig:agg_bpc_a}
\end{subfigure}
\vspace{5em}

\begin{subfigure}{.95\linewidth}
    \centering
\includegraphics[width=\linewidth]{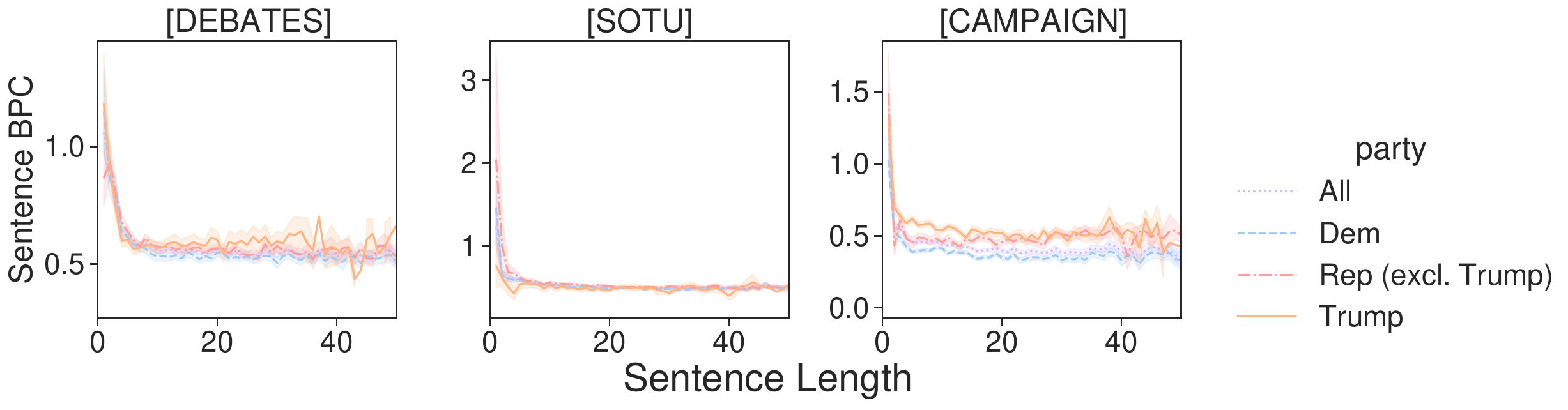}
\caption{Sentence BPC across sentence length}
\label{fig:agg_bpc_b}
\end{subfigure}
\caption{Average sentence BPC across each data type. In campaigns, Trump is the most unpredictable among those candidates; however, he is ranks closer to the middle for predictability in debates and SOTU }
\label{fig:agg_bpc}

\end{figure}

Recall that we employ bits-per-character (BPC), also known as bits-per-byte, as a measure of ``predictability'' of a sentence. Lower BPC values correspond to higher predictability. 
We use BPC instead of perplexity or loss directly to account for variation in tokenization techniques. 
BPC is the sum of cross-entropy losses of each token in a sentence, divided by the number of characters (bytes) in the sentences.
\figref{fig:agg_bpc} shows average sentence BPC per candidate (\figref{fig:agg_bpc_a}) and aggregated by party over sentence length (\figref{fig:agg_bpc_b}). While in campaigns, Trump is the most unpredictable among those candidates, he is ranks closer to the middle for predictability in debates and SOTU.

\subsection{Correlation with Readability Metrics}\label{apps:readability}
\begin{figure}
    \centering
    \begin{subfigure}{.45\linewidth}
    \includegraphics[width=\linewidth]{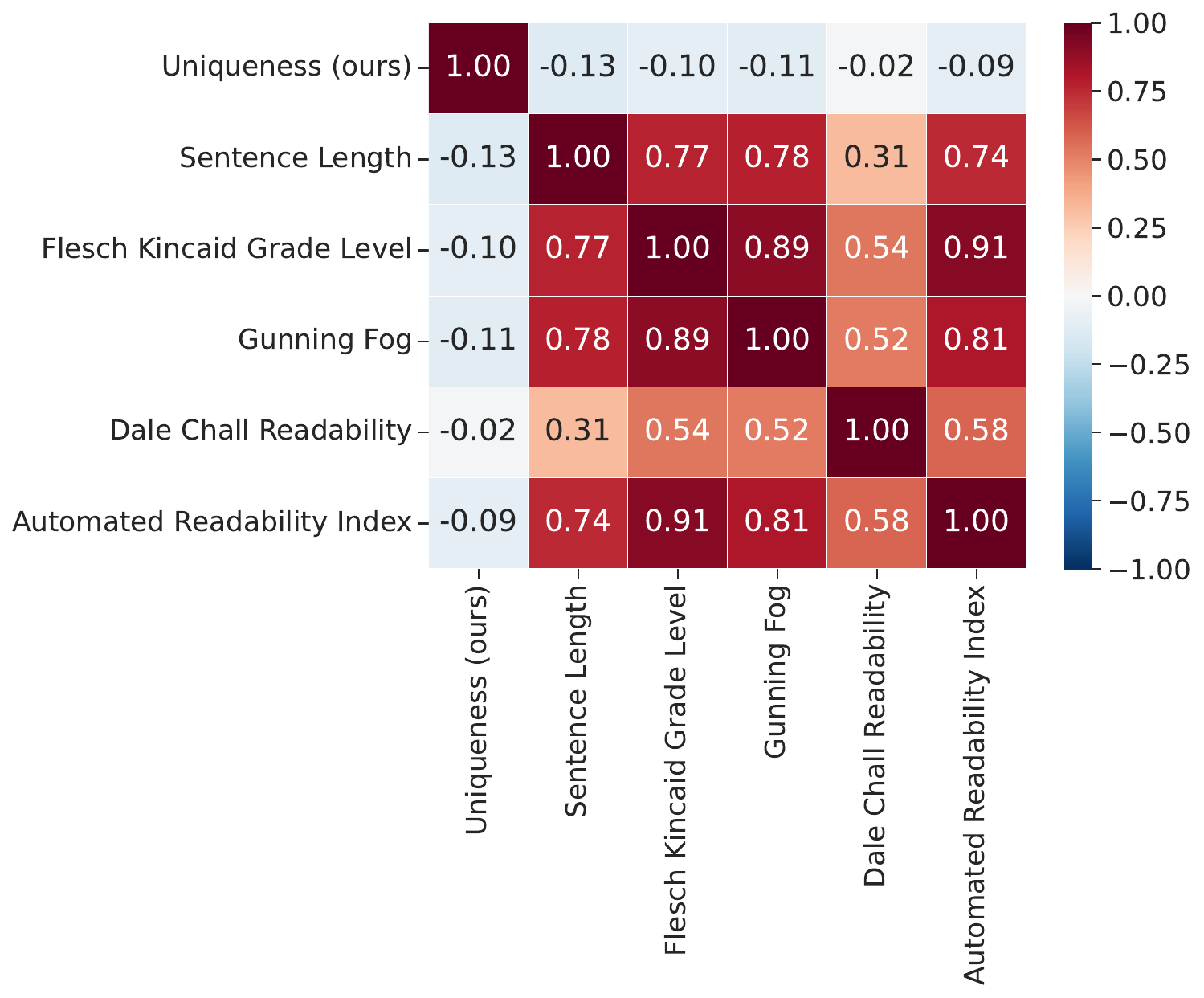}
    \caption{Debates}
    \label{fig:corr_debates}
    \end{subfigure}
    \hfill
    \begin{subfigure}{.45\linewidth}
    \includegraphics[width=\linewidth]{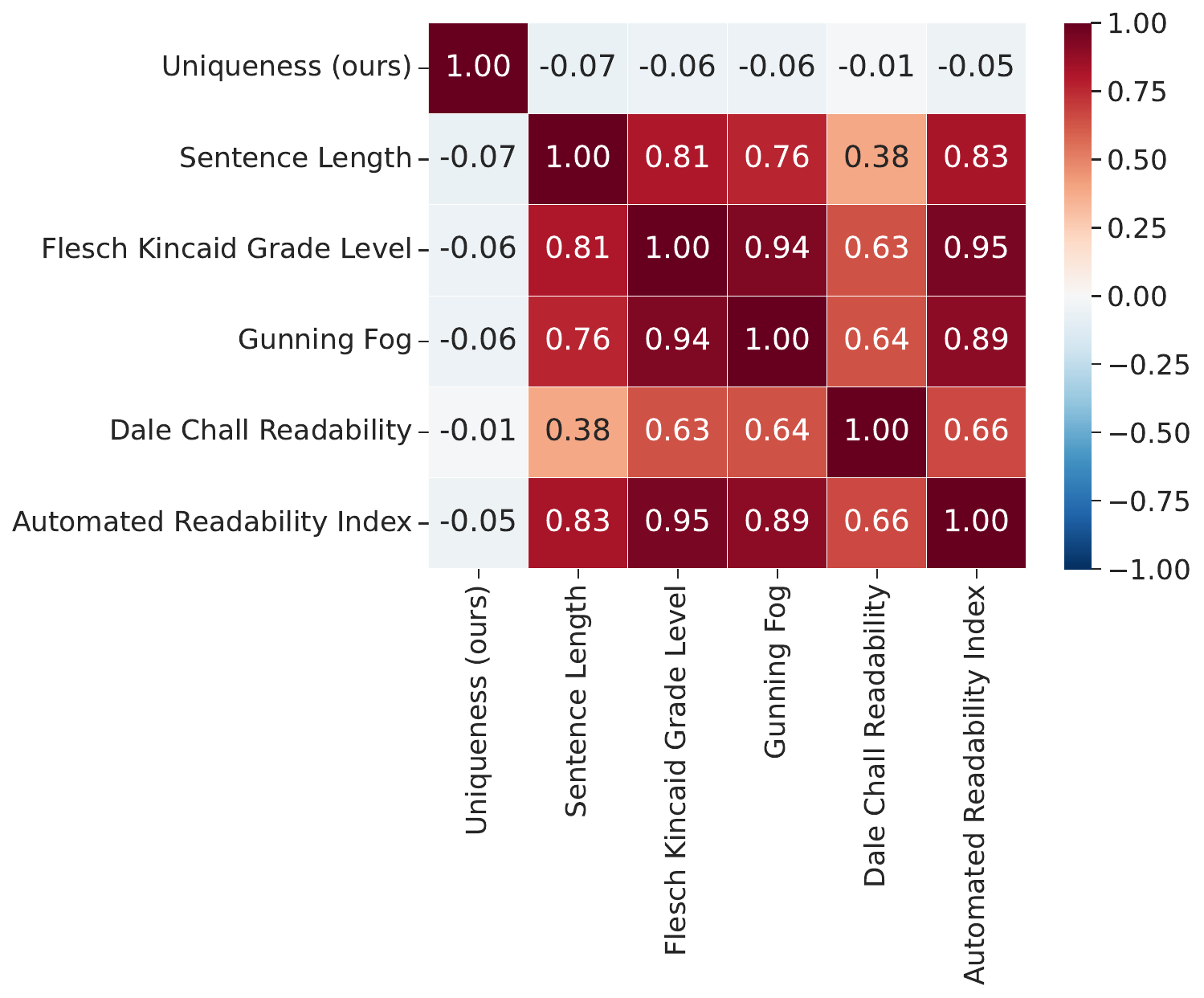}
    \caption{SOTU}
    \label{fig:corr_sotu}
    \end{subfigure}
\bigskip

    \begin{subfigure}{.45\linewidth}
    \includegraphics[width=\linewidth]{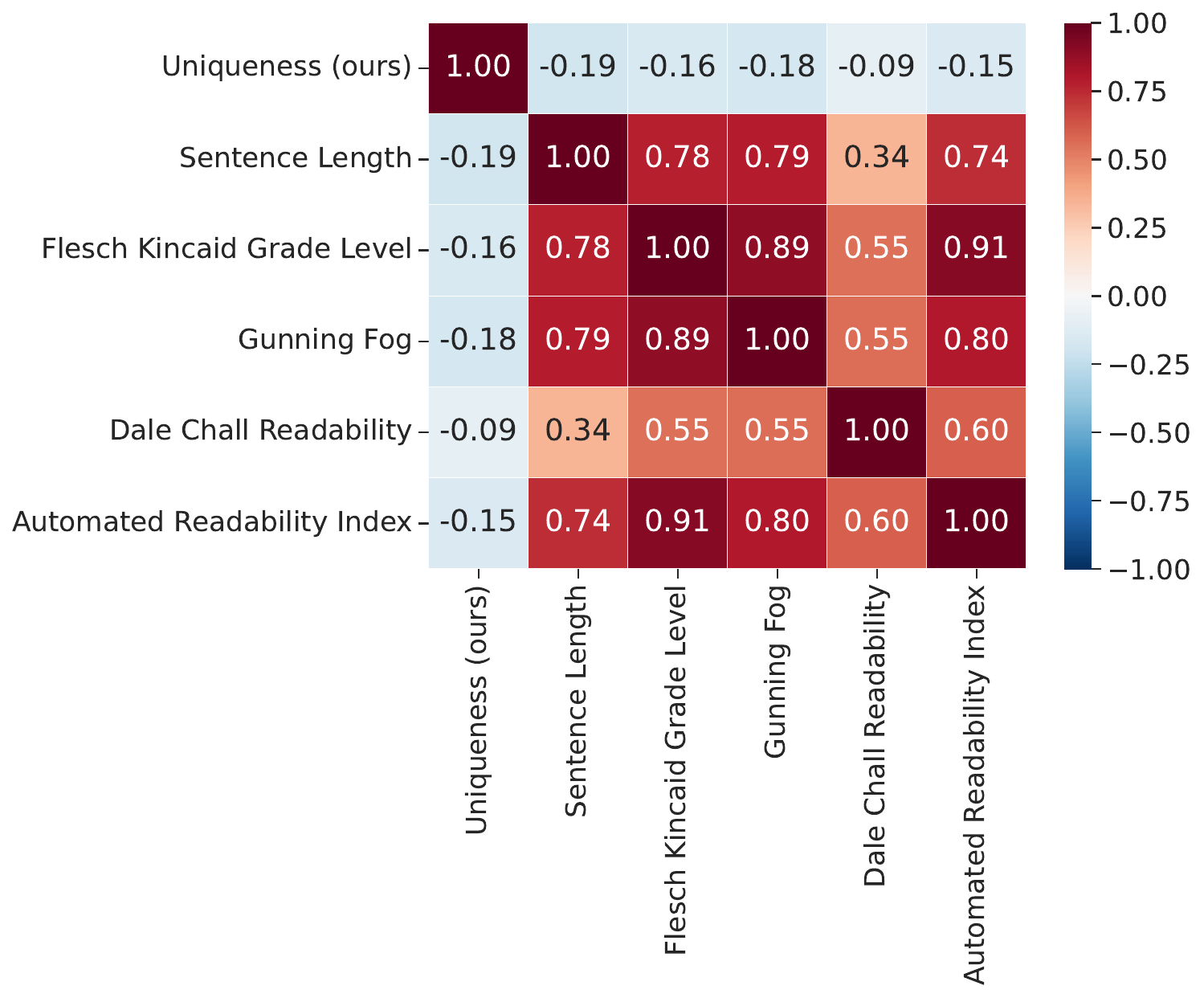}
    \caption{Campaign}
    \label{fig:corr_campaign}
        
    \end{subfigure}
    \caption{Heatmaps of Spearman Correlation Coefficients between our uniqueness metric and existing readability scores (FKGL, Gunning Fog, Dale Chall,  and ARI). There is little to no correlation between our score and readability; in contrast, there is strong correlation between the different readability indexes.}
    \label{fig:corr_maps}
    
\end{figure}

We calculate the Spearman correlation coefficients for our uniqueness metric with existing readability scores. Our metric has little to no correlation with readability, while different readability indexes show strong correlation with each other (\figref{fig:corr_maps}). \figref{fig:corr_maps} also confirms that our metric is dintinct from sentence length.

\subsection{Uniqueness scores over time}\label{apps:time_trends} 
\begin{figure}
    \centering
    \includegraphics[width=.95\linewidth]{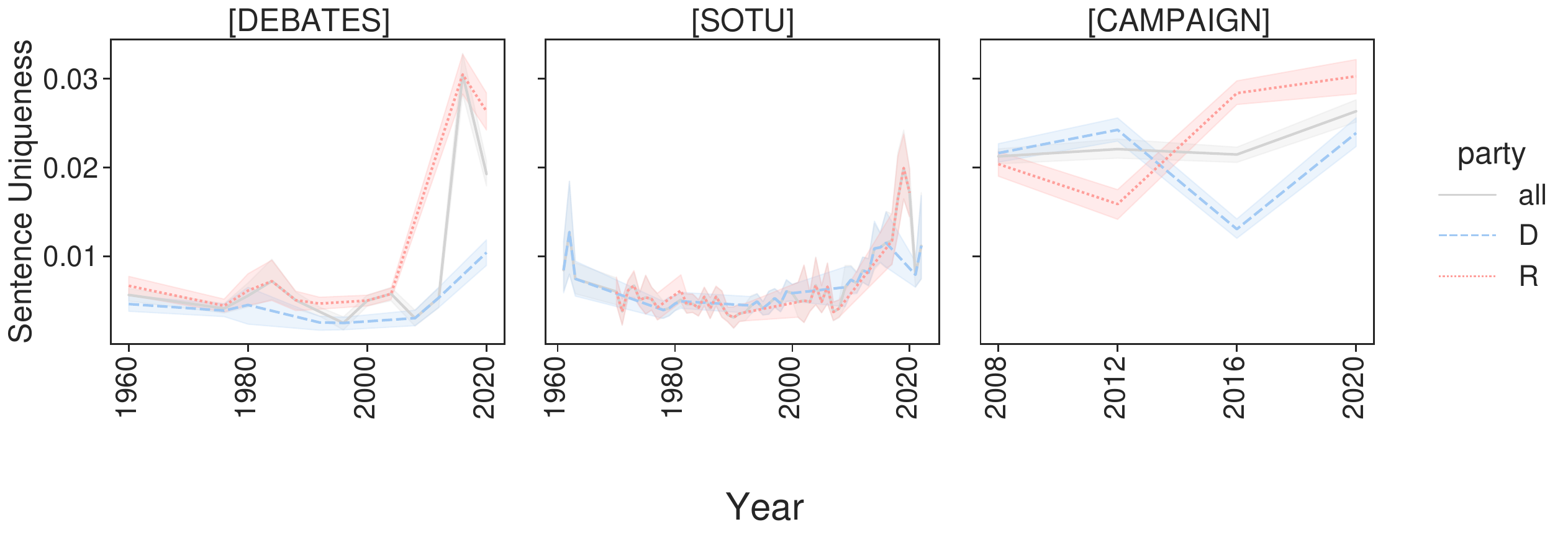}
    \caption{Uniqueness score over time, aggregated by political party.}
    \label{fig:uniq_over_time}
\end{figure}

\begin{figure*}%
\centering
\begin{subfigure}{0.45\linewidth}
    \centering
\includegraphics[width=.85\linewidth]{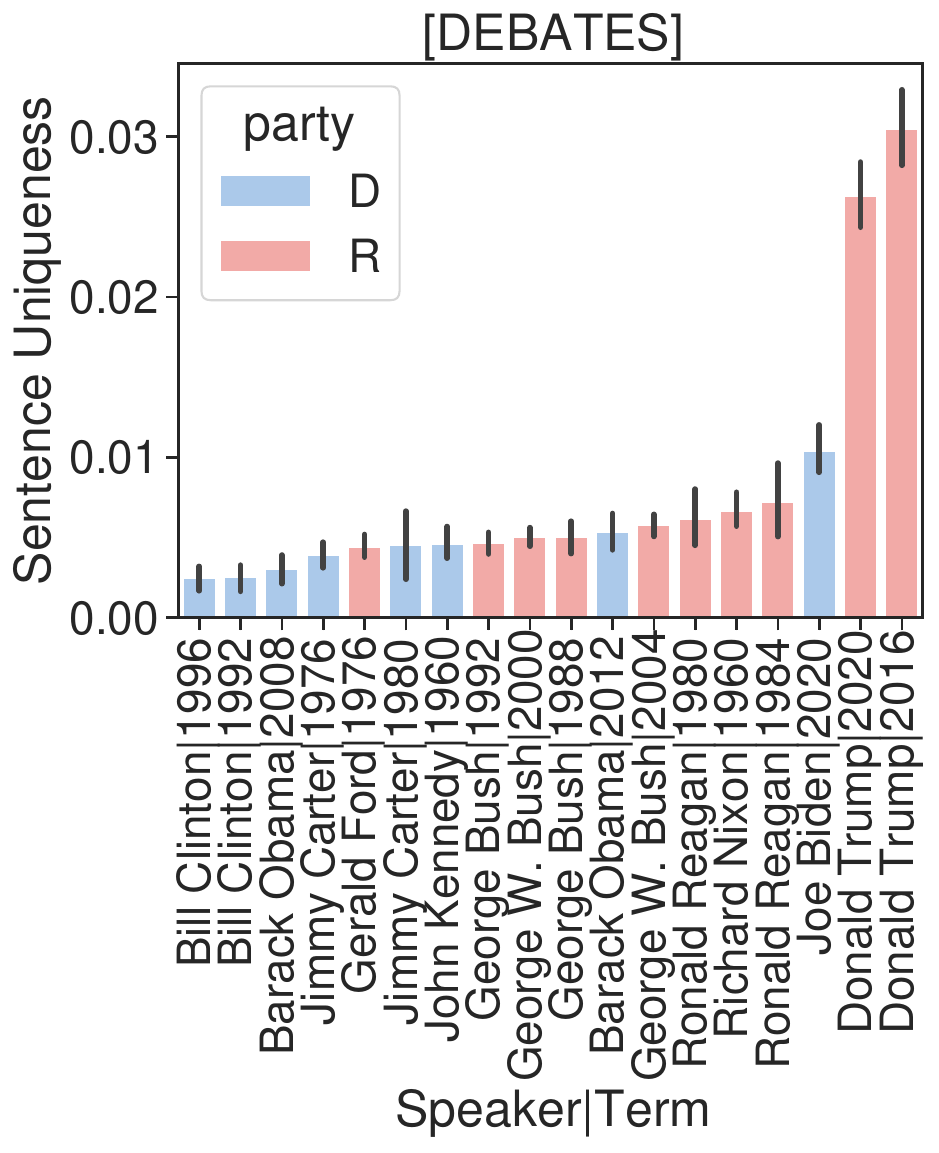}
\caption{Debates by election year}
\label{fig:debates_term}
\end{subfigure}
\begin{subfigure}{0.45\linewidth}
    \centering
    \includegraphics[width=.85\linewidth]{{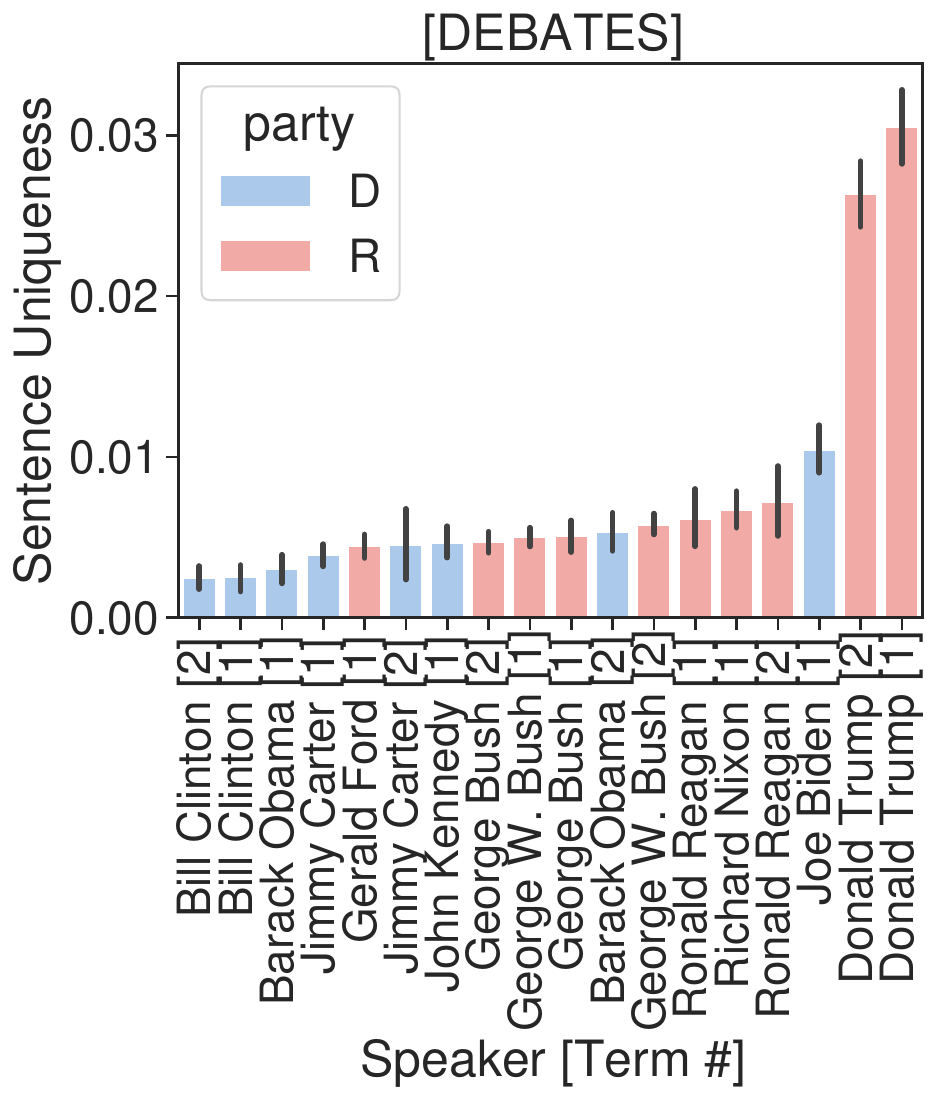}}
\caption{\label{fig:debates_order} Debates by term \#}
\end{subfigure}
\caption{\label{fig:debates_terms} For debates, Trump is consistently most unique for each term that he runs for president (2016, 2020). He is slightly more unique in his first election cycle, in 2016. The error bars represent 95\%-confidence intervals.}
\end{figure*}

\begin{figure*}%
\centering
\begin{subfigure}{0.45\linewidth}
    \centering
\includegraphics[width=.85\linewidth]{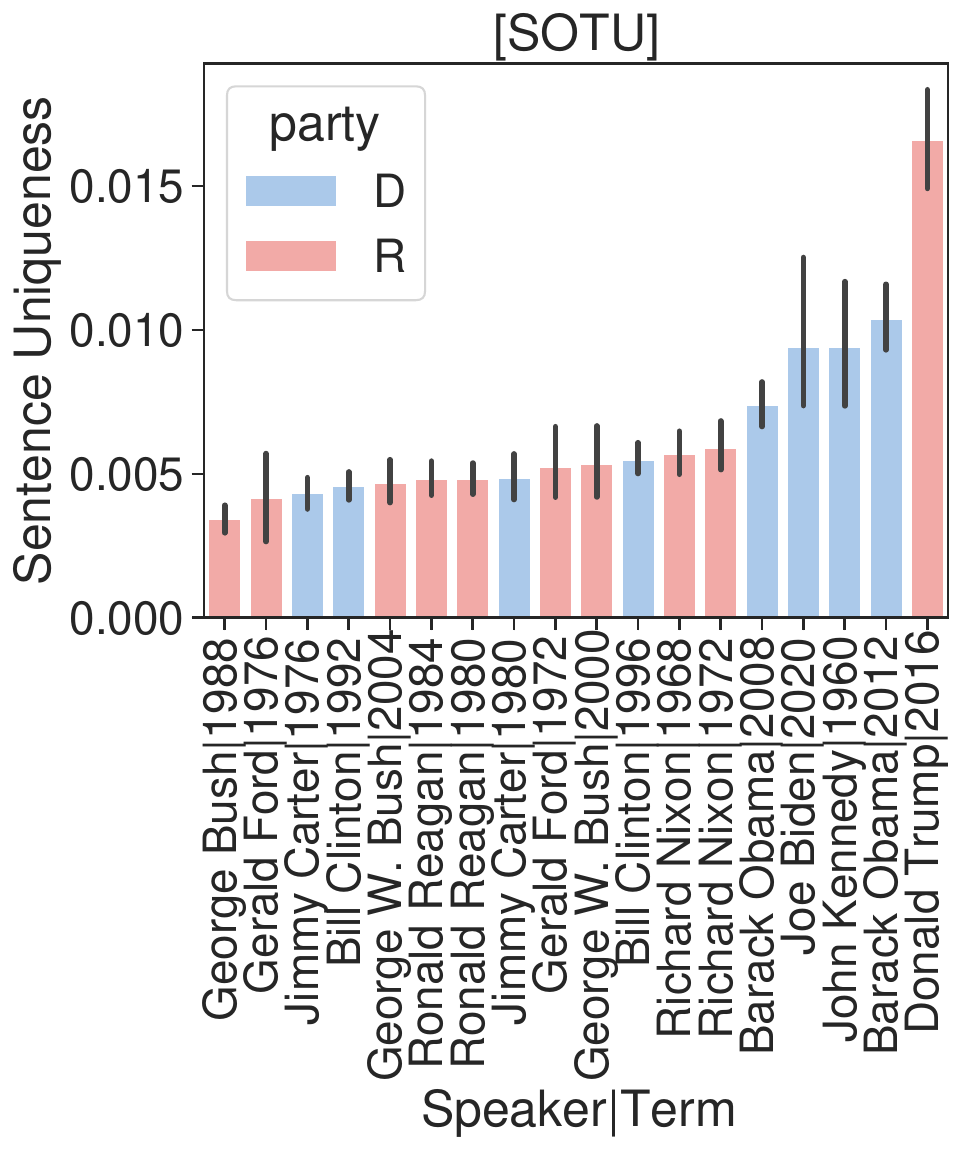}
\caption{SOTU by term}
\label{fig:sotu_term}
\end{subfigure}
\begin{subfigure}{0.45\linewidth}
    \centering
    \includegraphics[width=.85\linewidth]{{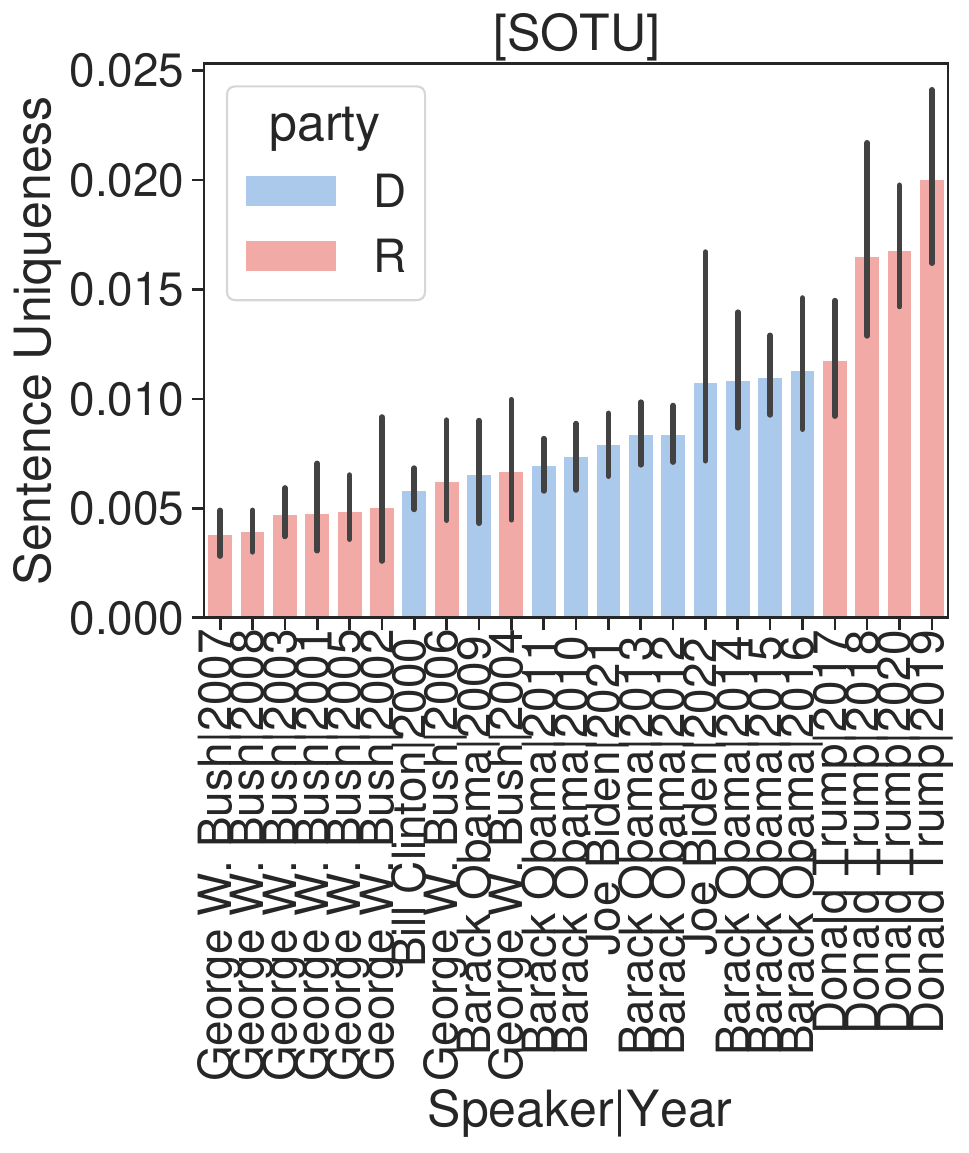}}
\caption{\label{fig:sotu_yr} SOTU by year}
\end{subfigure}
\caption{\label{fig:sotu_terms} For SOTU, Trump is consistently most unique throughout all years of his presidency. His speech distinctiveness increases after the first year of his term. The error bars represent 95\%-confidence intervals.}
\end{figure*}

\begin{figure*}%
\centering
\begin{subfigure}{0.45\linewidth}
    \centering
\includegraphics[width=.85\linewidth]{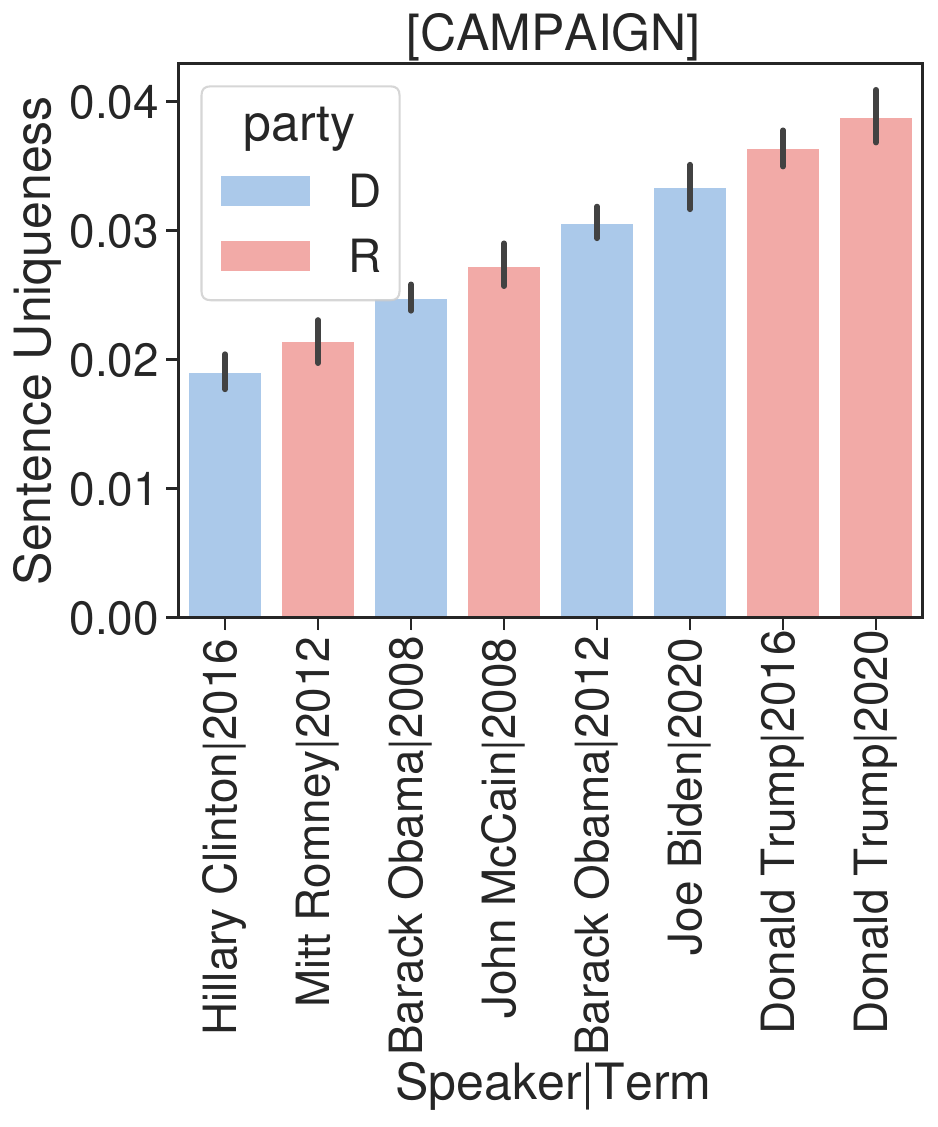}
\caption{Campaign by election year}
\label{fig:campaign_term}
\end{subfigure}
\begin{subfigure}{0.45\linewidth}
    \centering
    \includegraphics[width=.85\linewidth]{{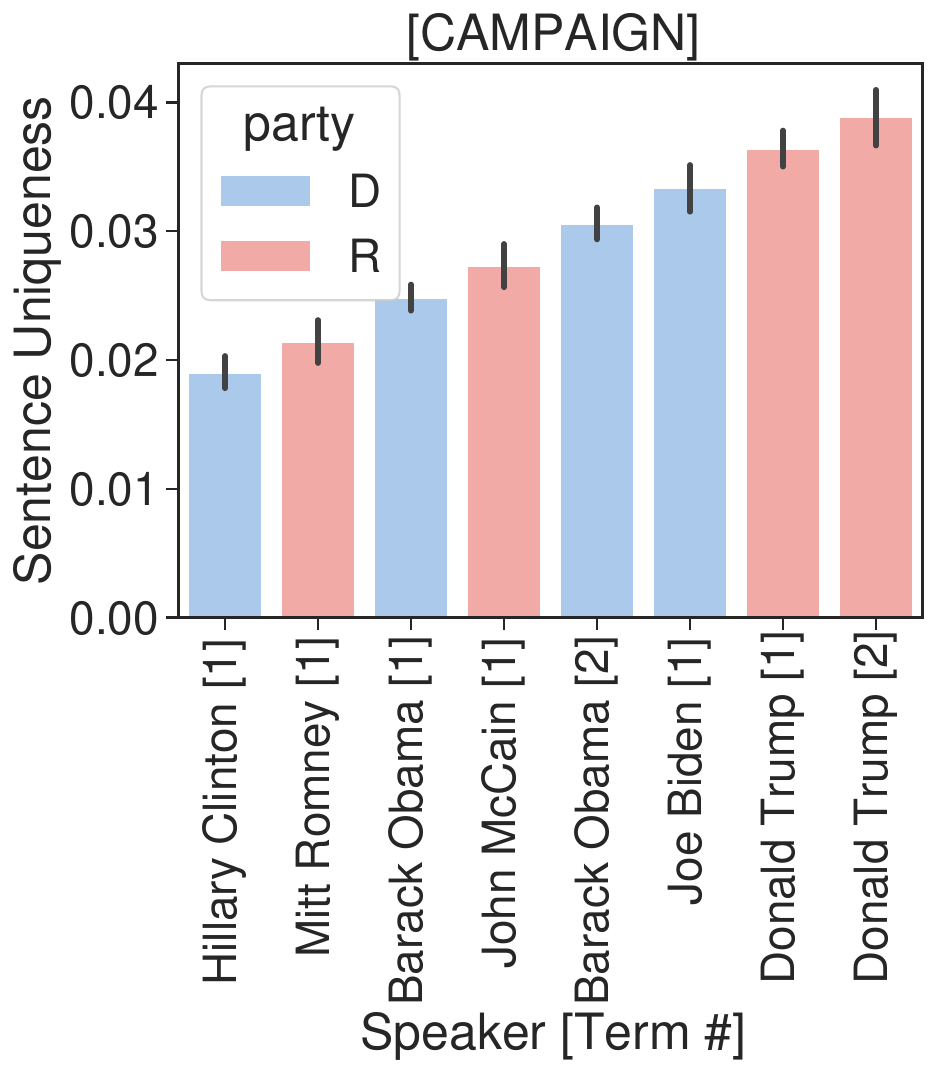}}
\caption{\label{fig:campaign_order} Campaign by term \#}
\end{subfigure}
\caption{\label{fig:campaign_terms} For campaign speeches, Trump is consistently most unique for each term that he runs for president (2016, 2020). His campaign speech in his second election cycle is slightly more unique than his first time. The error bars represent 95\%-confidence intervals.}
\end{figure*}

\figref{fig:uniq_over_time} shows the uniqueness scores over time; in particular, the scores increased in recent years. 
We also present the uniqueness scores by year/term for debates (\figref{fig:debates_terms}),
SOTU (\figref{fig:sotu_terms}), and
campaigns (\figref{fig:campaign_terms}).
Overall, Trump is still consistently the most unique every year and term, compared to those of the other presidential candidates. 
For debates, he is slightly more unique in his first election cycle in 2016, whereas in campaigns, his second election cycle involves more distinctive speech than his first. For SOTU addresses, Trump's speech becomes more distinct in his second year onwards.

\subsection{Top \%-ile uniqueness}\label{apps:decile}
\begin{figure}
    \centering
    \includegraphics[width=.95\linewidth]{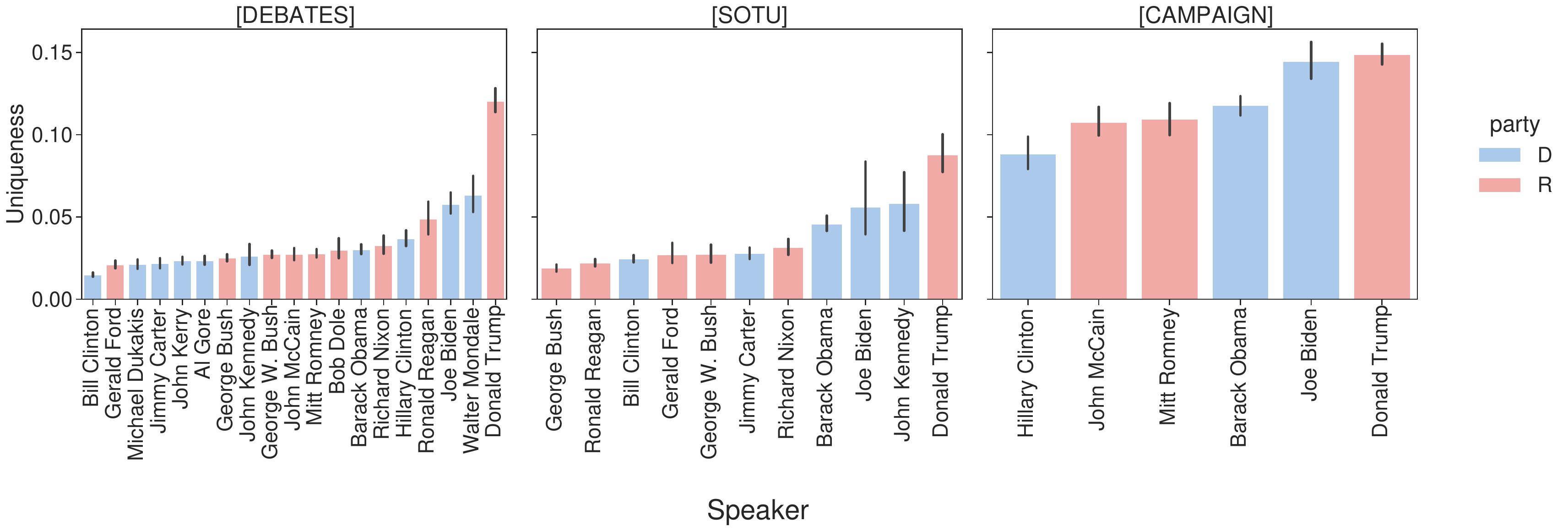}
    \caption{Uniqueness score by candidate, of sentences within the top 10-ile of scores for each candidate (the error bars represent 95\%-confidence intervals). Note that Trump continues to be most unique in all datasets.}
    \label{fig:uniq_decile}
\end{figure}

\figref{fig:uniq_decile} shows that, when comparing the uniqueness scores of the top decile of unique sentences for each speaker, we again find that Trump is the most unique speaker in all our samples of political speech.

\subsection{LLM-based uniqueness results with unmasked model}\label{apps:unmasked_model}
In the main paper, we present results using language models that are trained on data with named entities masked out with a \texttt{<ENT>} token, as identified by the spaCy NER tagger. This decision is made to prevent the model from learning patterns like ``Only Trump mentions Biden's name during debates''. 

\begin{figure*}%
\centering
\begin{subfigure}[t]{.95\linewidth}
    \centering
    \includegraphics[width=\linewidth]{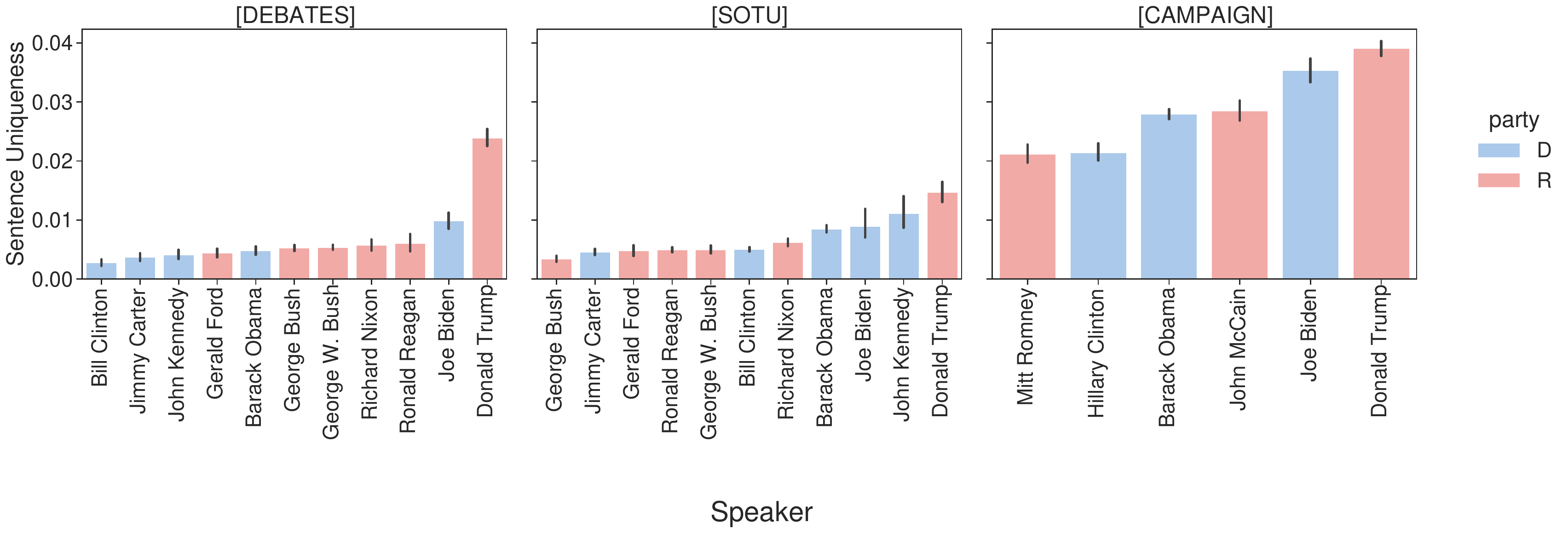}
\caption{Overall score by candidate}
\label{fig:unmasked_a}
\end{subfigure}
\vspace{5em}

\begin{subfigure}[t]{.95\linewidth}
    \centering
    \includegraphics[width=\linewidth]{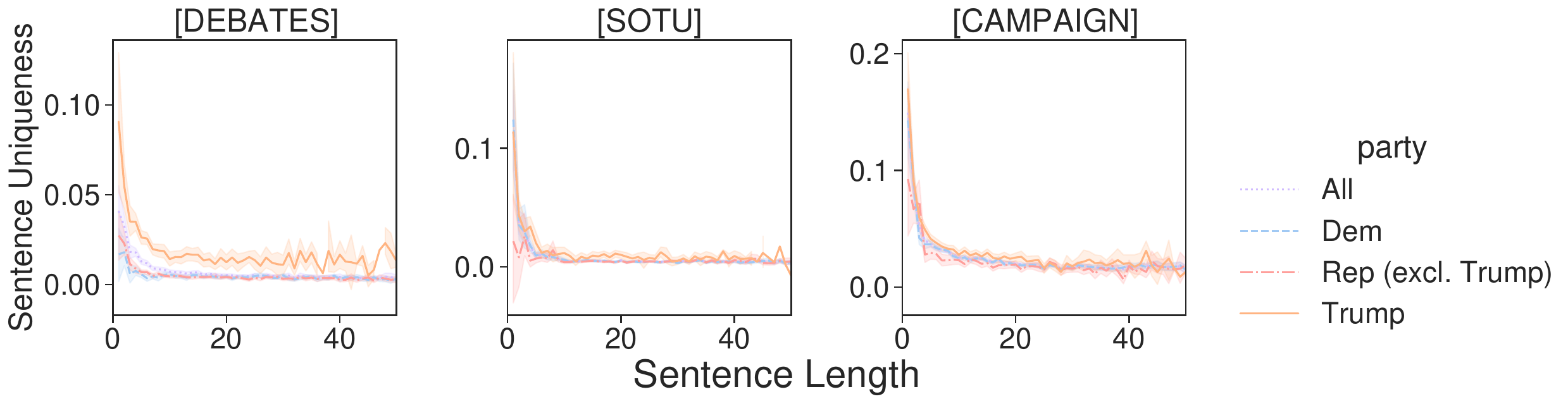}
\caption{Scores by sentence length}
\label{fig:unmasked_b}
\end{subfigure}
\vspace{5em}

\begin{subfigure}[t]{.95\linewidth}
    \centering
    \includegraphics[width=0.5\linewidth]{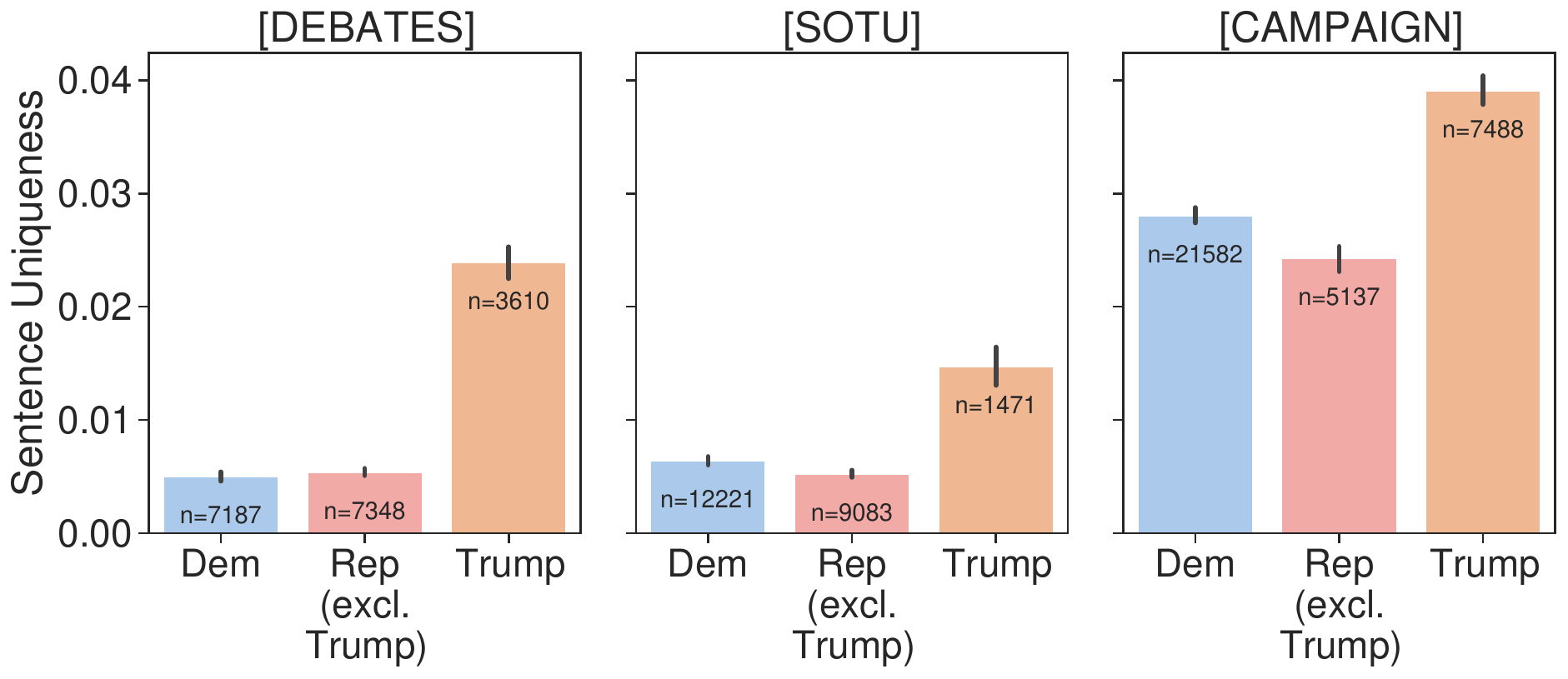}
    \caption{Uniqueness by party}
\label{fig:unmasked_c}
\end{subfigure}
\caption{Average sentence uniqueness for each speaker, across all data types, using the UNMASKED model (the error bars represent 95\%-confidence intervals). Trump is still the most unique speaker in all types of speeches.}
\label{fig:unmasked_agg_uniq_overall}
\end{figure*}

We also fine-tune each dataset's model on the data without masking named entities, with results presented in \figref{fig:unmasked_agg_uniq_overall}. These results are overall consistent with those of the masked model. \figref{fig:unmasked_a} shows that Trump remains the most distinctive speaker in all data types, while \figref{fig:unmasked_b} confirms still that the distinctiveness holds across all sentence lengths. Trump is still more similar in uniqueness to Democrats than his fellow Republicans (\figref{fig:unmasked_c}).

\begin{figure*}%
\centering
\begin{subfigure}[t]{.95\linewidth}
    \centering
    \includegraphics[width=.65\linewidth]{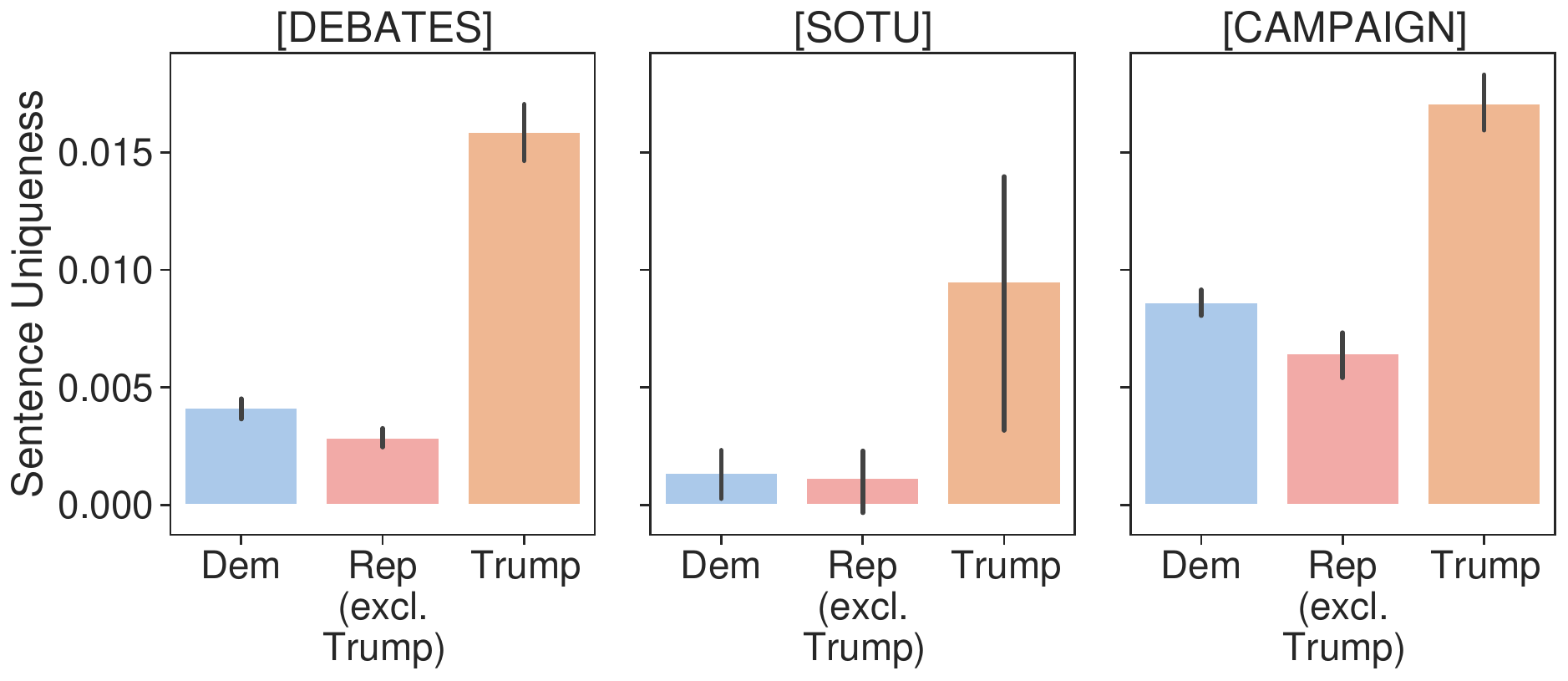}
    
    \caption{Gemma 2B}
\label{fig:valid_a}
\end{subfigure}
\vspace{3em}

\begin{subfigure}[t]{.95\linewidth}
    \centering
    \includegraphics[width=.65\linewidth]{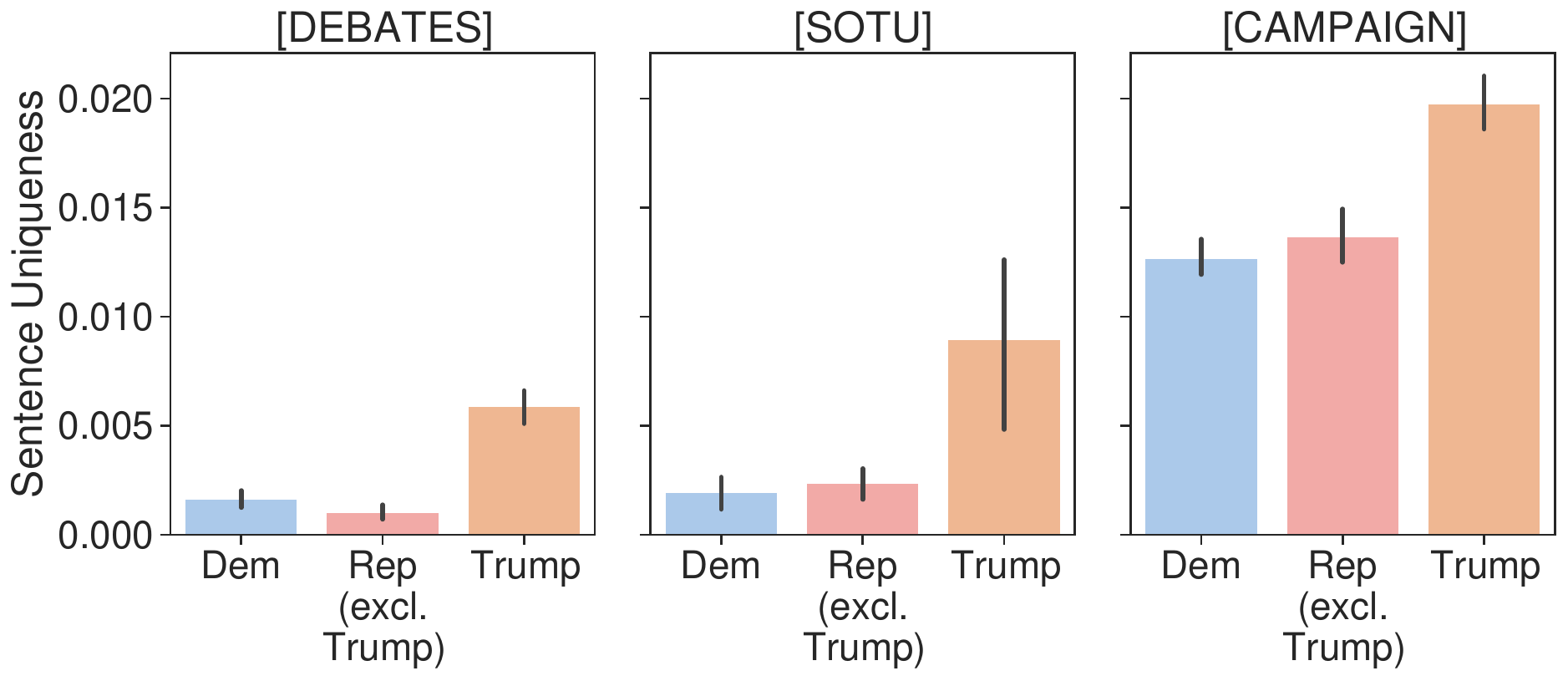}
    
    \caption{Phi-1.5}
\label{fig:valid_b}
\end{subfigure}

\caption{{\revision Validation of the LLM-based uniqueness metric with alternative LLMs. The resulting trends are consistent with original findings that Trump's speech is more unique than other Democratic and Republican presidential candidates.}}
\label{fig:valid_llm}
\end{figure*}

\begin{revision}

\subsection{Validation with other LLMs}\label{apps:valid_llm} We validate our uniqueness metric implementation with other LLMs. Specifically, we recalculate scores with the Gemma-2B and Phi-1.5 (1.3B) base models \cite{team2024gemma, textbooks2}. These models are considerably larger than the 124M parameters of GPT-2 \cite{radford2019language}.
Like our original setup up for GPT-2, we further train each model on all available corresponding data with a learning rate of $5e$-$5$. Each model is trained with LoRA tuning for 15 epochs, using LLaMA-Factory \cite{zheng2024llamafactory}.
As shown in \figref{fig:valid_llm}, while the exact scores differ, the trends are consistent with the main results: that is, Trump's speech is the most unique among the selection of Democratic and Republican presidential candidates.

\end{revision}

\section{Additional Divisive Word Lexicon Results}\label{app:div_results}

\subsection{Divisive Word Lexicon}\label{app:div_lexicon} 
\tbref{tab:div_lex_words} contains the 178 words in our proposed divisive word lexicon.

\subsection{Usage over time}\label{app:time_trends}
\begin{figure}
    \centering
     \centering
    \includegraphics[width=.85\linewidth]{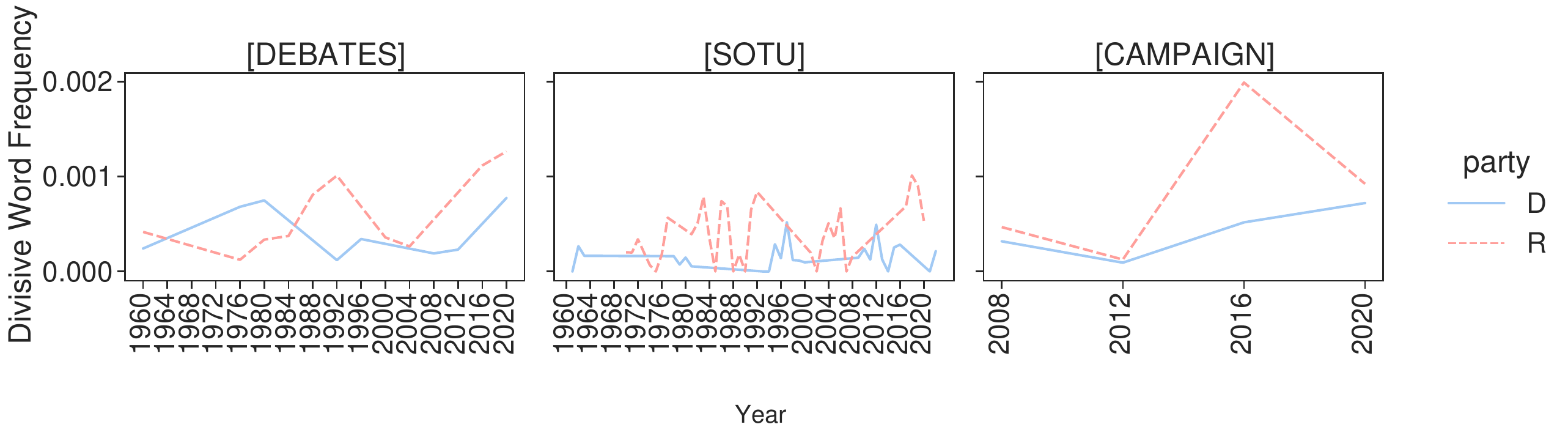}
\caption{\label{fig:lex_over_time} Frequency of usage over time}
\end{figure}

In debates and campaigns, divisive language usage increased after 2012 (\figref{fig:lex_over_time}), which corresponds to the onset of Trump's candidacy. Indeed, we find that Trump uses the most divisive language of all candidates.

\subsection{Usage by candidate}
\begin{figure*}
  \centering
  
  \begin{subfigure}{0.95\linewidth}
    \centering
    \includegraphics[width=\linewidth]{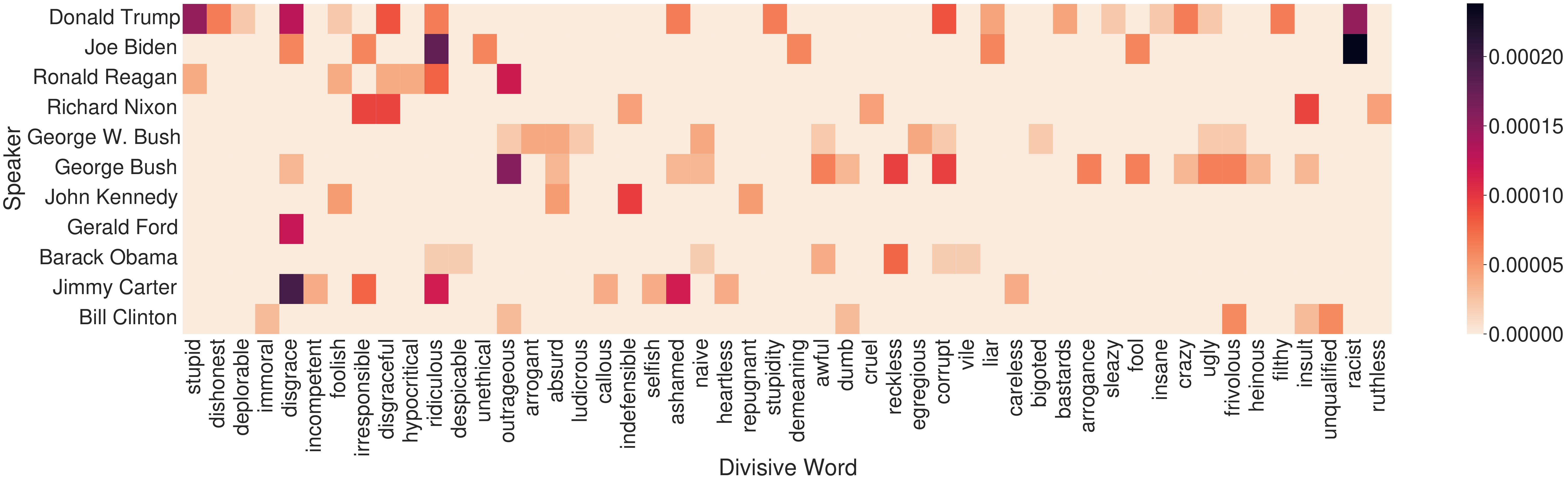}
    \caption{Debates}
    \label{fig:subfigureA}
  \end{subfigure}
  \vspace{3em}
  
  \begin{subfigure}{0.95\linewidth}
    \centering
    \includegraphics[width=\linewidth]{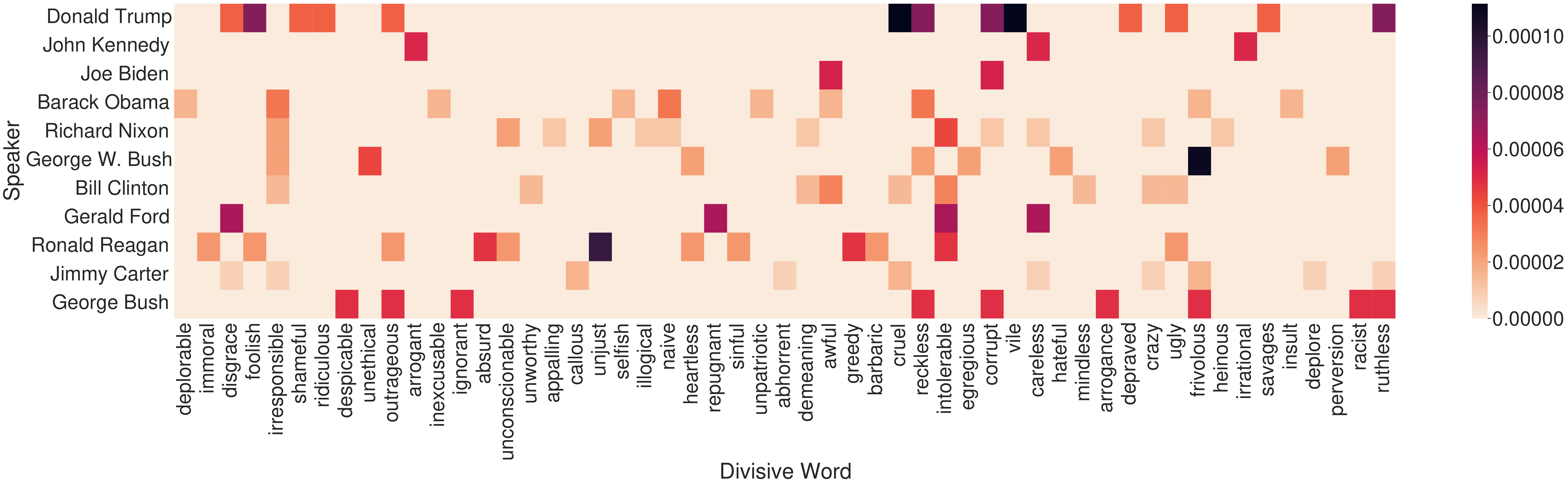}
    \caption{SOTU}
    \label{fig:subfigureB}
  \end{subfigure}
  \vspace{3em}
  
  \begin{subfigure}{0.95\linewidth}
    \centering
    \includegraphics[width=\linewidth]{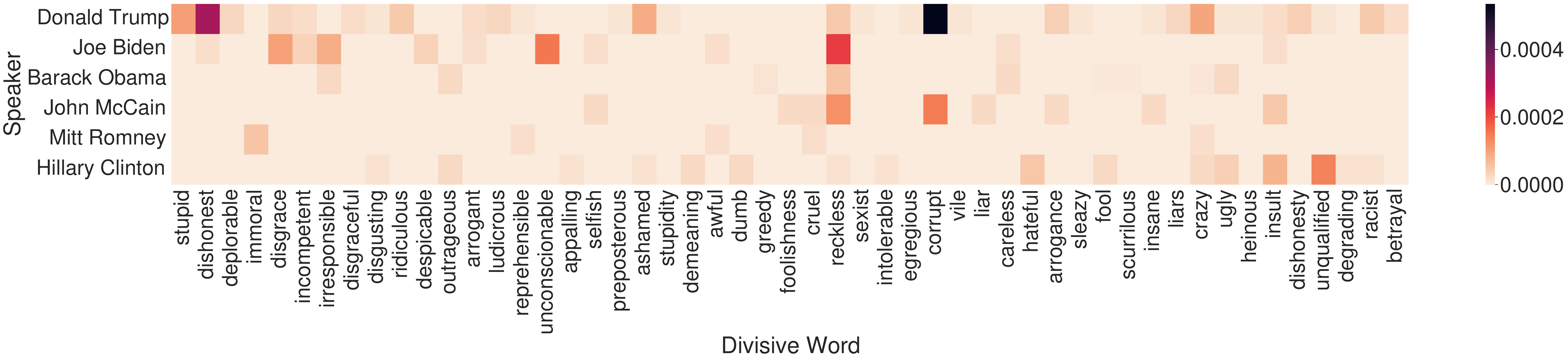}
    \caption{Campaign}
    \label{fig:subfigureC}
  \end{subfigure}
  
  \caption{Heatmaps showing divisive word utterance frequency, normalized over all words spoken, by speaker. }
  \label{fig:lex_main}
\end{figure*}

Heatmaps of divisive word usage can provide a more granular look at the specific divisive language used by each candidate (\figref{fig:lex_main}).
For example, in the debates dataset, it is notable that the frequency of use of the word “racist” has become used more by recent candidates like Trump and Biden. 
In the debates dataset some of Donald Trump's most frequently used words in this lexicon are those like ``disgrace'', ``stupid'', ``filthy'', ``hate'', and ``racist'' (\figref{fig:subfigureA}).
In Donald Trump's SOTU addresses, the most frequently used divisive words are those like ``corrupt'', ``vile'', ``foolish'', ``cruel'', and ``savage'' (\figref{fig:subfigureB}).
In campaign speeches, Donald Trump's most frequently used divisive words are those like ``corrupt'', ``crazy'', ``stupid'', and ``dishonest'' (\figref{fig:subfigureC}).

\subsection{Utterances Using Divisive Words}\label{app:div_excerpts} Excerpts of speech using divisive words can be found in \tbref{tab:debates_div_ex}, \tbref{tab:sotu_div_ex}, and \tbref{tab:campaign_div_ex} for debates, SOTU, and campaign speeches respectively.

\begin{revision}
\subsection{Annotator Agreement}\label{app:annot_agreement}
For the initial 360 terms, we obtain a Fleiss' $\kappa$ of $0.54$, which indicates a moderate level of agreement. Since divisiveness is a somewhat subjective phenomenon, this moderate agreement is similar to that obtained on other subjective lexicons, such as the word-emotion association  (average Fleiss' $\kappa$ of 0.29) \cite{mohammad2013crowdsourcing} and the initial terms in the polarization dictionary (Cohen's $\kappa$ of 0.61)  \cite{10.1093/pnasnexus/pgac019}. Our final set of 178 terms has full agreement from the majority of annotators, but we still discuss the implications of our Fleiss' $\kappa$ below.

Agreement is affected by several characteristics of divisiveness. 
Firstly , divisiveness is a spectrum and terms can be ``a little'' to ``extremely'' divisive. Annotators only give binary labels, so disagreement may be higher due to lack of nuance. 
Relatedly, the words are presented out of context; they may be more or less divisive depending on usage. However, identifying all possible contexts of our lexicon usage is costly and nontrivial.
Finally, like the word-emotion association lexicon~\cite{mohammad2013crowdsourcing}, our initial data is biased, with more ``not divisive'' terms than ``divisive'' terms (based on only 31.6\% of the 360 terms receiving unanimous agreement). Both Cohen's and Fleiss' $\kappa$ are considered conservative in such settings with label imbalance~\cite{mohammad2013crowdsourcing, Brennan_Prediger_1981, perreault_reliability-1989}. As such, these $\kappa$ values may underestimate the true rater agreement for these cases.

Ultimately, we introduce the first such lexicon of divisiveness for text analysis. Usage of our lexicon may be augmented with additional strategies like dependency parsing and leveraging surrounding context.

\end{revision}

\section{Additional Opponent Mention Results}\label{app:opp_results}

\subsection{Rate of opponent mentions} 
\begin{figure*}%
    \centering
    \includegraphics[width=.85\linewidth]{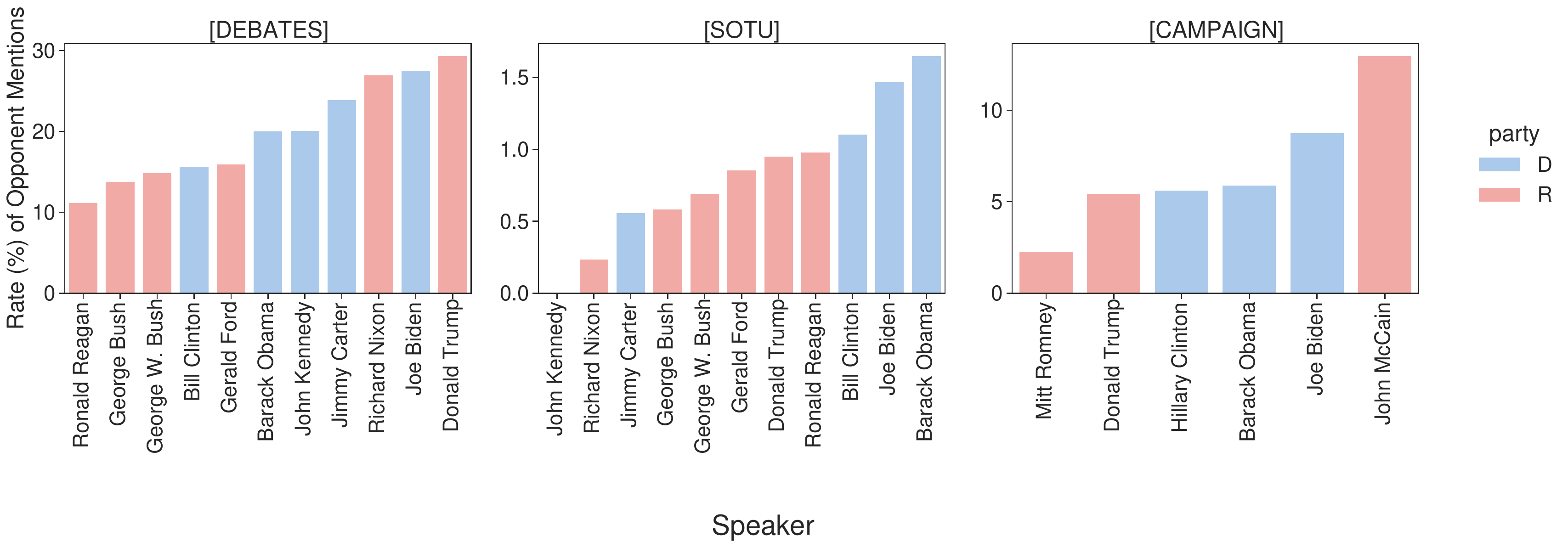}
\caption{Distribution of sentences containing opponent mentions among all speakers, across the data types. Note the different rates for debates vs. SOTU. Trump has the highest rate of opponent mentions in debates. }
\label{fig:opp_rate_ment}
\end{figure*}

The overall rates of sentences that mention opponents for debates, SOTU, and campaign speeches are 20.60\%,  0.83\%, and 6.95\% respectively. \figref{fig:opp_rate_ment} shows the distribution of sentences containing opponent mentions among the candidate speakers. Trump has the highest rate of opponent mentions in debates.

\subsection{Fightin' Words overlap}\label{app:fw_overlap}
\begin{figure}%
\centering
\begin{subfigure}{.75\linewidth}
    \centering
    \includegraphics[width=\linewidth]{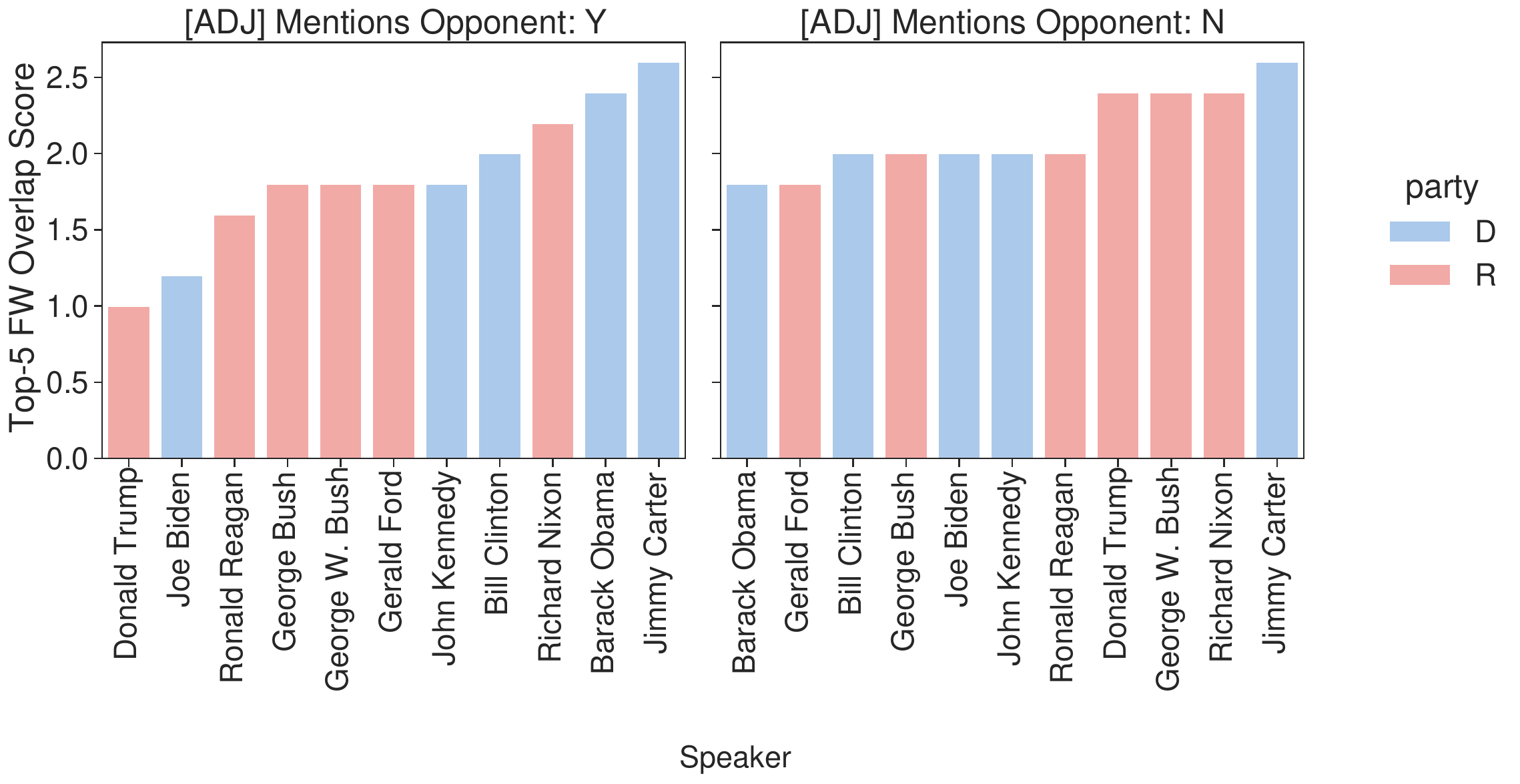}
    \caption{\label{fig:fw_a} Top-5 FW overlap score in debates for each candidate}
\end{subfigure}
\medskip

\begin{subfigure}{.75\linewidth}
    \centering
    \includegraphics[width=\linewidth]{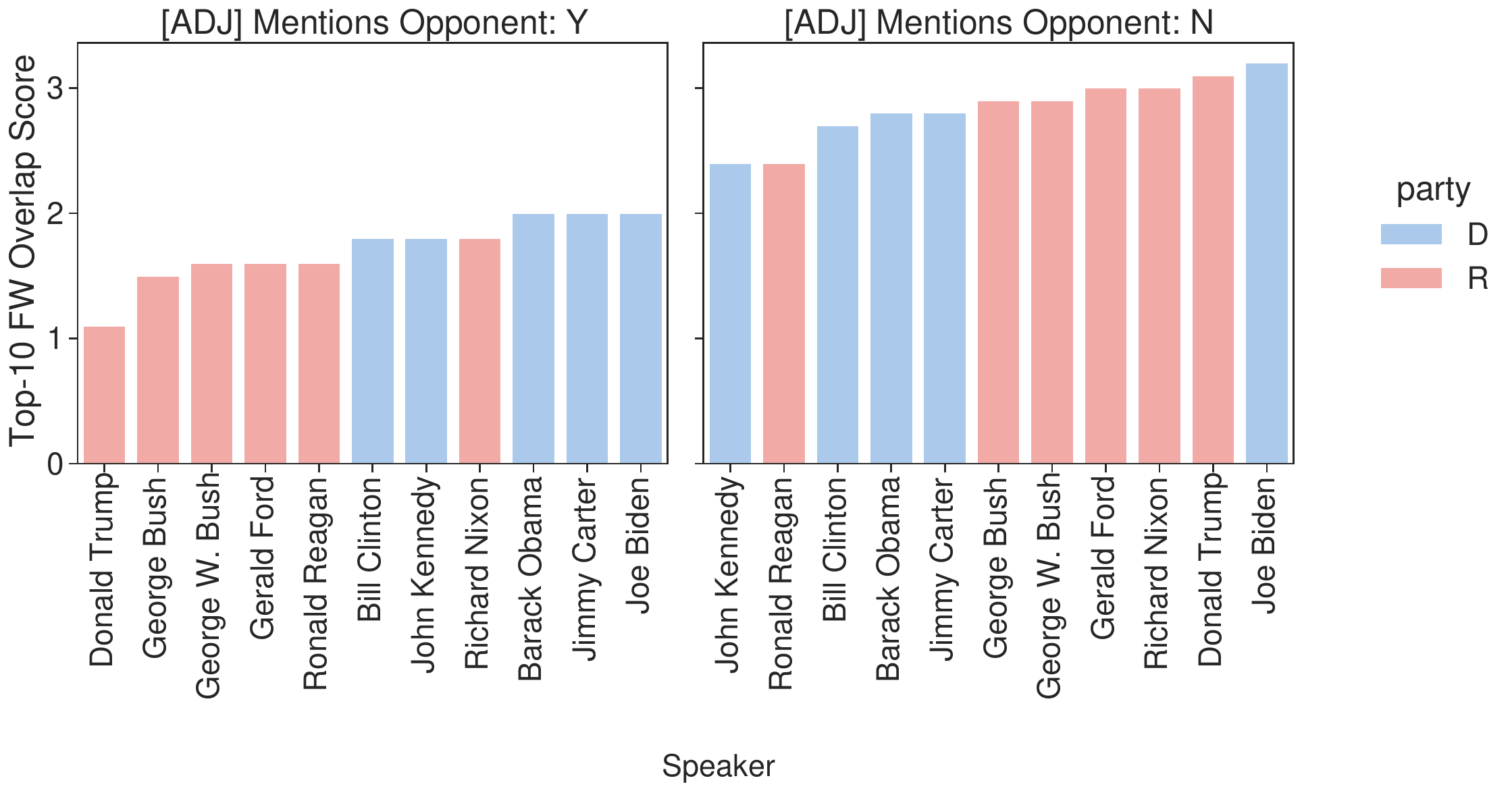}
    \caption{\label{fig:fw_b} Top-10 FW overlap score in debates for each candidate}
\end{subfigure}
\medskip

\begin{subfigure}{.75\linewidth}
    \centering
    \includegraphics[width=\linewidth]{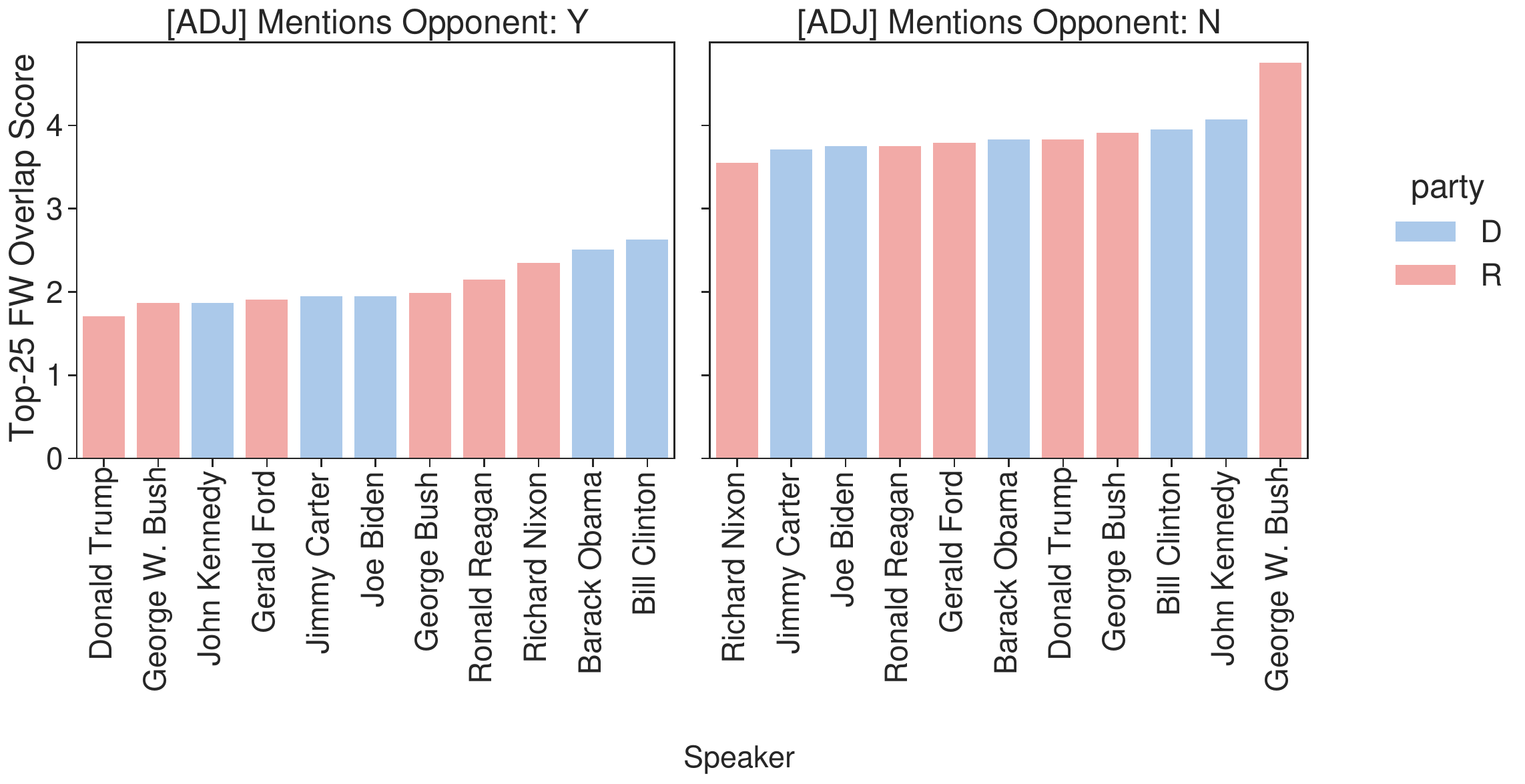}
    \caption{\label{fig:fw_b} Top-25 FW overlap score in debates for each candidate}
\end{subfigure}
\caption{\label{fig:fw_plots} Fightin' Words overlap results from debates. Trump's FW associated with opponent mentions generally have the lowest overlap in adjective usage compared to other candidates, which is another indication that his language is distinctive. 
}
\end{figure}

We present additional plots for the Fightin' Words overlap metric, for different top-$n$ adjectives in debates (see \figref{fig:fw_plots}). In particular, for top-$5$, $10$, and $25$ adjective Fightin' Words, Trump has the lowest overlap in adjectives he uses in association with opponent mentions. These trends are consistent with those presented in the main paper.

\begin{figure}
    \centering
    
\begin{subfigure}{.75\linewidth}
    \includegraphics[width=\linewidth]{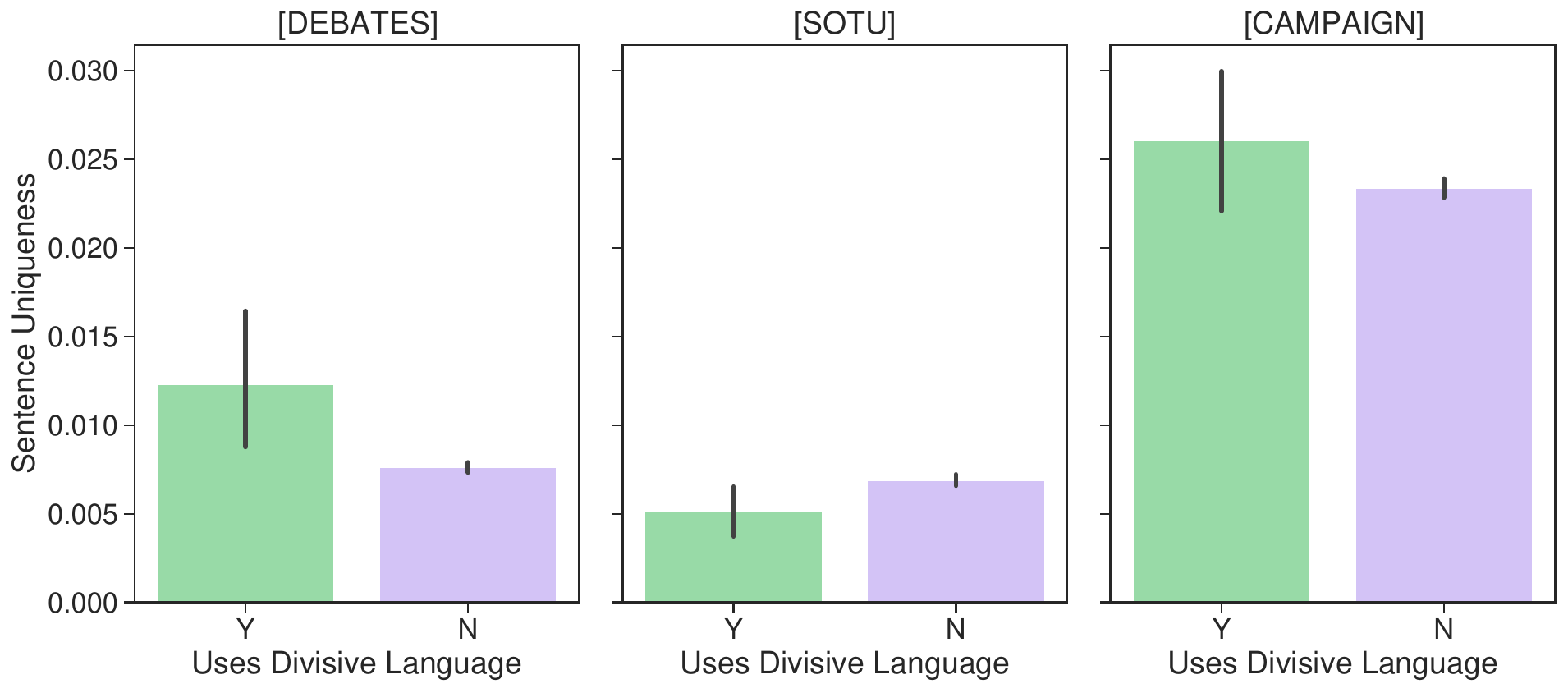}
    \caption{\revisionTwo Sentence uniqueness across divisive language usage, for each type of speech. }
    \label{fig:comp_1}
\end{subfigure}
\bigskip

\begin{subfigure}{.35\linewidth}
    \includegraphics[width=\linewidth]{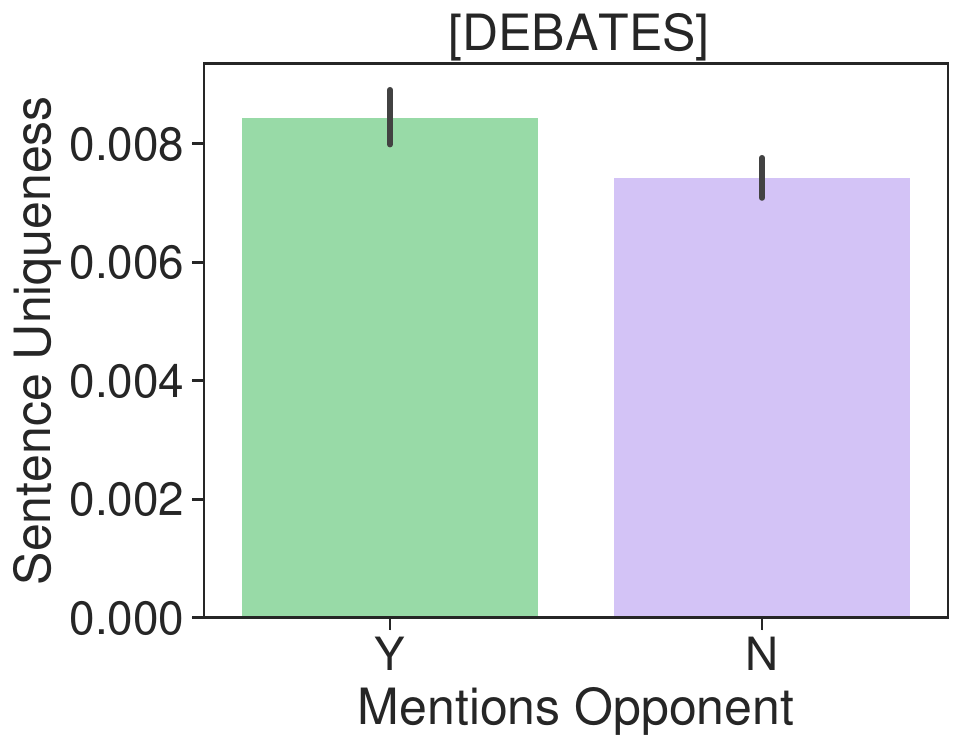}
    \caption{\revisionTwo Sentence uniqueness across opponent mentions, in debates.}
    \label{fig:comp_2}
\end{subfigure}
\hspace{5em}
\begin{subfigure}{.35\linewidth}
    \includegraphics[width=\linewidth]{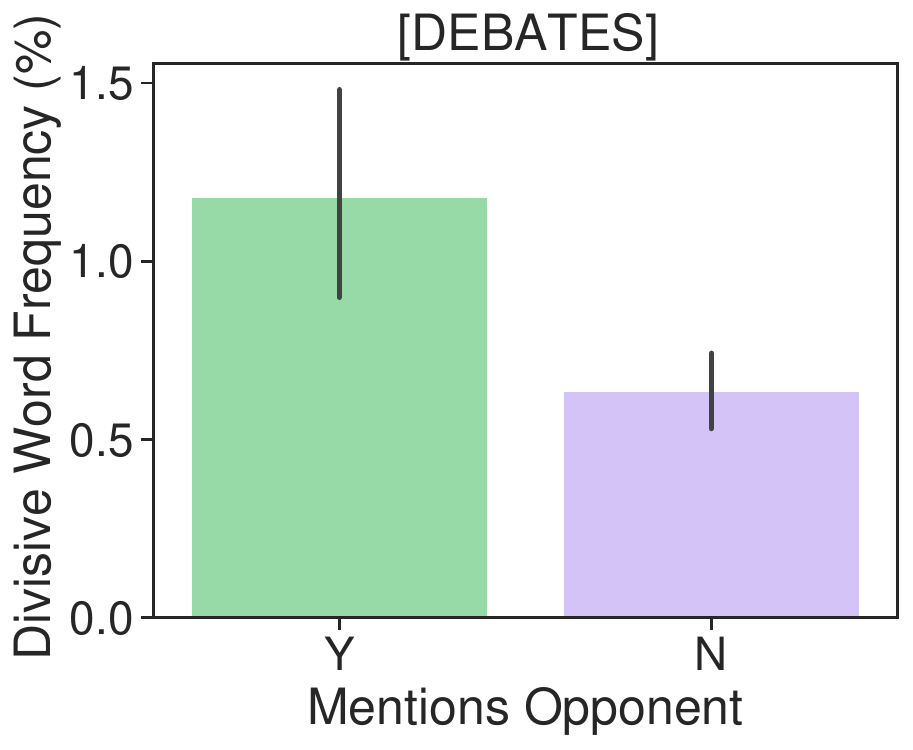}
    \caption{\revisionTwo Divisive language frequency across opponent mentions, in debates}
    \label{fig:comp_3}
\end{subfigure}

    \caption{\revisionTwo Pairwise analysis of our three proposed metrics, aggregated globally across all speakers. \figref{fig:comp_1} shows that in debates and campaign speeches, sentences that use divisive language tend to be more unique as well. For sentences containing opponent mentions in debates, \figref{fig:comp_2} shows that such utterances tend to be more distinctive and \figref{fig:comp_3} shows that they tend to have higher divisive word usage as well.}
    \label{fig:comp_metrics}
    
\end{figure}

\begin{revisionTwo}
\section{Relating uniqueness, divisiveness, and opponent mentions.} \label{app:comp_metrics}

\figref{fig:comp_1} shows that in debates and campaign speeches, sentences that use divisive language tend to be more unique as well. The Spearman correlation coefficient between divisiveness and uniqueness is 0.01 for debates and campaigns ($p<$0.05 in campaigns and $p=$0.30 in debates).
For sentences containing opponent mentions in debates, \figref{fig:comp_2} shows that such utterances tend to be more distinctive and \figref{fig:comp_3} shows that they tend to have higher divisive word usage as well. The Spearman correlation for opponent mentions is 0.05 and 0.02 ($p < $0.05) between uniqueness and divisiveness, respectively.

\end{revisionTwo}

\begin{table}
    \centering
    \begin{tabular}{p{2.5cm} p{2.5cm} p{2.5cm} p{2.5cm} }
    \toprule
  stupid$^*$, \newline dishonest$^*$, \newline unamerican$^*$, \newline idiot$^*$, \newline deplorable$^*$, \newline pathetic$^*$, \newline immoral$^*$, \newline disgrace$^*$, \newline incompetent$^*$, \newline foolish$^*$,\newline irresponsible, \newline shameful, \newline disgraceful, \newline disgusting, \newline hypocritical, \newline idiotic, \newline unprofessional, \newline ridiculous, \newline despicable, \newline unethical, \newline outrageous, \newline arrogant, \newline inexcusable, \newline ignorant, \newline shameless, \newline disrespectful, \newline absurd, \newline ludicrous, \newline reprehensible, \newline unconscionable, \newline scandalous, \newline contemptible, \newline inept, \newline unworthy, \newline appalling, \newline laughable, \newline disingenuous, \newline cowardly, \newline callous, \newline unjust, \newline indefensible, \newline foolhardy, \newline selfish, \newline preposterous, \newline deceitful 
 & 
 
disloyal, \newline ashamed, \newline illogical, \newline distasteful, \newline naive, \newline heartless, \newline crass, \newline repugnant, \newline sinful, \newline unpatriotic, \newline abhorrent, \newline obnoxious, \newline childish, \newline thoughtless, \newline stupidity, \newline demeaning, \newline untrustworthy, \newline awful, \newline egotistical, \newline deluded, \newline dumb, \newline greedy, \newline boorish, \newline barbaric, \newline tasteless, \newline foolishness, \newline cruel, \newline amoral, \newline pretentious, \newline hypocrite, \newline revolting, \newline reckless, \newline manipulative, \newline nonsensical, \newline idiots, \newline sexist, \newline intolerable, \newline uncivilized, \newline egregious, \newline duplicitous, \newline undignified, \newline atrocious, \newline corrupt, \newline contemptuous, \newline vile 
 & 
 
liar, \newline vindictive, \newline careless, \newline wretched, \newline hateful, \newline bigoted, \newline bastards, \newline undemocratic, \newline feckless, \newline cowardice, \newline pompous, \newline mindless, \newline spineless, \newline arrogance, \newline rude, \newline treasonous, \newline vulgar, \newline disgusted, \newline insufferable, \newline sleazy, \newline fool, \newline perverse, \newline scurrilous, \newline insolent, \newline grotesque, \newline devious, \newline incompetence, \newline buffoon, \newline unintelligent, \newline undisciplined, \newline odious, \newline insane, \newline liars, \newline depraved, \newline crazy, \newline moronic, \newline uncouth, \newline petulant, \newline ugly, \newline elitist, \newline frivolous, \newline hypocrisy, \newline dastardly, \newline incapable, \newline tyrannical 
 & 
 
fools, \newline heinous, \newline thuggish, \newline impulsive, \newline detestable, \newline demented, \newline traitor, \newline hypocrites, \newline filthy, \newline conceited, \newline traitors, \newline loathsome, \newline barbarous, \newline irrational, \newline savages, \newline gullible, \newline obscene, \newline repulsive, \newline delusional, \newline insult, \newline scoundrel, \newline abominable, \newline deplore, \newline mockery, \newline perversion, \newline bumbling, \newline dopey, \newline inane, \newline nauseating, \newline brainless, \newline incorrigible, \newline exploitive, \newline gutless, \newline dishonesty, \newline unqualified, \newline conniving, \newline promiscuous, \newline degrading, \newline racist, \newline ruthless, \newline perverted, \newline diabolical, \newline betrayal 
 \\
        \bottomrule
    \end{tabular}
    \caption{The 178 words in our divisive word lexicon, which are verified as being divisive by the majority of 4 annotators. Our original ten seed words are denoted with a $^*$.}
    \label{tab:div_lex_words}
\end{table}

\begin{table}[]
    \centering
    \footnotesize
    \begin{tabular}{p{1cm} p{14cm}}
    \toprule

    \textbf{Year} &
    \textbf{Example Sentence(s) w/ Context} \\ 
\cmidrule(lr){1-2}
2020 & 
\textbf{Joe Biden:} I'm talking about the Biden plan. . . \newline
\textbf{Donald Trump:} Where they want to rip down buildings. . . \newline
\textbf{Other:} [to Biden] Let him go for a minute, and then you can go. \newline
\textbf{Donald Trump:} And rebuild the building. \newline
\textbf{Joe Biden:} No. \newline
\textbf{Donald Trump:} It's the \textbf{dumbest}- \newline
\textbf{Joe Biden:} That is not, that is not. . . \newline
\textbf{Donald Trump:} ... most \textbf{ridiculous}. . . \\ 
\cmidrule(lr){1-2}
2020 & 
\textbf{Joe Biden:} Russia is paying you a lot. \newline
\textbf{Joe Biden:} China is paying a lot. \newline
\textbf{Joe Biden:} And your hotels and all your businesses all around the country, all around the world. \newline
\textbf{Joe Biden:} And China's building a new road to a new gas  a golf course you have overseas. \newline
\textbf{Joe Biden:} So what's going on here? \newline
\textbf{Joe Biden:} Why don't you release your tax return or stop talking about \textbf{corruption}? \\ 
\cmidrule(lr){1-2}
2020 & 
\textbf{Donald Trump:} We have to go back to the core values of this country. \newline
\textbf{Donald Trump:} They were teaching people that our country is a horrible place. \newline
\textbf{Donald Trump:} It's a \textbf{racist} place. \newline
\textbf{Donald Trump:} And they were teaching people to hate our country. \newline
\textbf{Donald Trump:} And I'm not going to allow that to happen. \newline
\textbf{Joe Biden:} Nobody's doing that. \newline
\textbf{Other:} Vice President Biden. \newline
\textbf{Joe Biden:} Nobody's doing that. \newline
\textbf{Joe Biden:} He's the \textbf{racist}. \\ 
\cmidrule(lr){1-2}
2020 & 
\textbf{Donald Trump:} I mean, they can say anything. \newline
\textbf{Donald Trump:} It's a very-- it makes me sad because I am the least \textbf{racist} person. \newline
\textbf{Donald Trump:} I can't even see the audience because it's so dark, but I don't care who's in the audience. \newline
\textbf{Donald Trump:} I'm the least \textbf{racist} person in this room. \newline
\textbf{Other:} OK. \newline
\textbf{Other:} Vice President Biden, let me ask you, very quickly, and then I have a follow up question for you. \newline
\textbf{Joe Biden:} Abraham Lincoln. \newline
\textbf{Joe Biden:} Here is one of the most \textbf{racist} presidents we've had in modern history. \newline
\textbf{Joe Biden:} He pours fuel on every single \textbf{racist} fire, every single one. \\ 
 \cmidrule(lr){1-2}
      2016 &
\textbf{Donald Trump:} Well, all of these bad leaders from ISIS are leaving Mosul. \newline
\textbf{Donald Trump:} Why can't they do it quietly? \newline
\textbf{Donald Trump:} Why can't they do the attack, make it a sneak attack, and after the attack is made, inform the American public that we've knocked out the leaders, we've had a tremendous success? \newline
\textbf{Donald Trump:} People leave. \newline
\textbf{Donald Trump:} Why do they have to say we're going to be attacking Mosul within the next four to six weeks, which is what they're saying? \newline
\textbf{Donald Trump:} How \textbf{stupid} is our country? \\ 
\cmidrule(lr){1-2}
2016 & 
\textbf{Donald Trump:} But the leaders that we wanted to get are all gone because they're smart. \newline
\textbf{Donald Trump:} They say, what do we need this for? \newline
\textbf{Donald Trump:} So Mosul is going to be a wonderful thing. \newline
\textbf{Donald Trump:} And Iran should write us a letter of thank you, just like the really \textbf{stupid} the \textbf{stupidest} deal of all time, a deal that's going to give Iran absolutely nuclear weapons. \\ 
\cmidrule(lr){1-2}
2016 & 
\textbf{Hillary Clinton:} And we should demand that Donald release all of his tax returns so that people can see what are the entanglements and the financial relationships that he has... \newline
\textbf{Other:} We're going to get to that later. \newline
\textbf{Other:} Secretary Clinton, you're out of time. \newline
\textbf{Hillary Clinton:} ... with the Russians and other foreign powers. \newline
\textbf{Other:} Mr. Trump? \newline
\textbf{Donald Trump:} Well, I think I should respond, because so \textbf{ridiculous}. \\
\cmidrule(lr){1-2}
1976 & 
\textbf{Gerald Ford:} On the other hand, when you have a bill of that magnitude, with those many provisions, a President has to sit and decide if there is more good than bad. \newline
\textbf{Gerald Ford:} And from the analysis that I have made so far, it seems to me that that tax bill does justify my signature and my approval. \newline
\textbf{Other:} Governor Carter, your response. \newline
\textbf{Jimmy Carter:} Well, Mr. Ford is changing considerably his previous philosophy. \newline
\textbf{Jimmy Carter:} The present tax structure is a \textbf{disgrace} to this country. \\ 
\cmidrule(lr){1-2}
1988 & 
\textbf{George Bush:} In terms of negative campaigning, you know, I don't want to sound like a kid in the schoolyard: he started it. \newline
\textbf{George Bush:} But take a look at the Democratic convention take a look at it. \newline
\textbf{George Bush:} Do you remember the Senator from Boston chanting out there and the ridicule factor from that lady from Texas that was on there; I mean, come on, this was just \textbf{outrageous}. \\ 
    \bottomrule
    \end{tabular}
    \caption{[DEBATES] Example usages of words from our divisive lexicon throughout debates. }
    \label{tab:debates_div_ex}
\end{table}

\begin{table}[]
    \centering
    \footnotesize
    \begin{tabular}{p{1cm} p{14cm}}
    \toprule
    \textbf{Year} &
    \textbf{Example Sentence(s) w/ Context} \\ 
    \cmidrule(lr){1-2}
    2022 & 
\textbf{Joe Biden:} Putin is now isolated from the world more than he has ever been. \newline
\textbf{Joe Biden:} Together, along with our allies, we are right now enforcing powerful economic sanctions. \newline
\textbf{Joe Biden:} We're cutting off Russia's largest banks from the international financial system; preventing Russia's Central Bank from defending the Russian ruble, making Putin's \$630 billion war fund worthless. \newline
\textbf{Joe Biden:} We're choking Russia's access to technology that will sap its economic strength and weaken its military for years to come. \newline
\textbf{Joe Biden:} Tonight I say to the Russian oligarchs and the \textbf{corrupt} leaders who have bilked billions of dollars off this violent regime: No more. \\ 
\cmidrule(lr){1-2}
    2020 & 
\textbf{Donald Trump:} The terrorist responsible for killing Sergeant Hake was Qasem Soleimani, who provided the deadly roadside bomb that took Chris's life. \newline
\textbf{Donald Trump:} Soleimani was the Iranian regime's most \textbf{ruthless} butcher, a monster who murdered or wounded thousands of American servicemembers in Iraq. \\ 
\cmidrule(lr){1-2}
2019 & 
\textbf{Donald Trump:} On Friday, it was announced that we added another 304,000 jobs last month alone, almost double the number expected. \newline
\textbf{Donald Trump:} An economic miracle is taking place in the United States, and the only thing that can stop it are \textbf{foolish} wars, politics, or ridiculous, partisan investigations. \\ 
\cmidrule(lr){1-2}
    2018 & 
\textbf{Donald Trump:} I am asking Congress to address the fundamental flaws in the terrible Iran nuclear deal. \newline
\textbf{Donald Trump:} My administration has also imposed tough sanctions on the communist and socialist dictatorships in Cuba and Venezuela. \newline
\textbf{Donald Trump:} But no regime has oppressed its own citizens more totally or brutally than the \textbf{cruel} dictatorship in North Korea. \\ 
\cmidrule(lr){1-2}
    2018 & 
\textbf{Donald Trump:} In April, this will be the last time you will ever file under the old and very broken system, and millions of Americans will have more take-home pay starting next month—a lot more. \newline
\textbf{Donald Trump:} We eliminated an especially \textbf{cruel} tax that fell mostly on Americans making less than \$50,000 a year, forcing them to pay tremendous penalties simply because they couldn't afford Government-ordered health plans. \\ 
\cmidrule(lr){1-2}
2017 & 
\textbf{Donald Trump:} We cannot allow a beachhead of terrorism to form inside America. \newline
\textbf{Donald Trump:} We cannot allow our Nation to become a sanctuary for extremists. \newline
\textbf{Donald Trump:} That is why my administration has been working on improved vetting procedures, and we will shortly take new steps to keep our Nation safe and to keep those out who will do us harm. \newline
\textbf{Donald Trump:} As promised, I directed the Department of Defense to develop a plan to demolish and destroy ISIS, a network of lawless savages that have slaughtered Muslims and Christians, and men and women and children of all faiths and all beliefs. \newline
\textbf{Donald Trump:} We will work with our allies, including our friends and allies in the Muslim world, to extinguish this \textbf{vile} enemy from our planet. \\ 
\cmidrule(lr){1-2}
2016 & 
\textbf{Barack Obama:} But after years now of record corporate profits, working families won't get more opportunity or bigger paychecks just by letting big banks or big oil or hedge funds make their own rules at everybody else's expense. \newline
\textbf{Barack Obama:} Middle class families are not going to feel more secure because we allowed attacks on collective bargaining to go unanswered. \newline
\textbf{Barack Obama:} Food stamp recipients did not cause the financial crisis; \textbf{recklessness} on Wall Street did. \\ 
\cmidrule(lr){1-2}
2012 & 
\textbf{Barack Obama:} In 2008, the house of cards collapsed. \newline
\textbf{Barack Obama:} We learned that mortgages had been sold to people who couldn't afford or understand them. \newline
\textbf{Barack Obama:} Banks had made huge bets and bonuses with other people's money. \newline
\textbf{Barack Obama:} Regulators had looked the other way or didn't have the authority to stop the bad behavior. \newline
\textbf{Barack Obama:} It was wrong, it was \textbf{irresponsible}, and it plunged our economy into a crisis that put millions out of work, saddled us with more debt, and left innocent, hard-working Americans holding the bag. \\ 
\cmidrule(lr){1-2}
1991 & 
\textbf{George Bush:} Last year, our friends and allies provided the bulk of the economic costs of Desert Shield. \newline
\textbf{George Bush:} And now, having received commitments of over \$40 billion for the first 3 months of 1991, I am confident they will do no less as we move through Desert Storm. \newline
\textbf{George Bush:} But the world has to wonder what the dictator of Iraq is thinking. \newline
\textbf{George Bush:} If he thinks that by targeting innocent civilians in Israel and Saudi Arabia, that he will gain advantage, he is dead wrong. \newline
\textbf{George Bush:} If he thinks that he will advance his cause through tragic and \textbf{despicable} environmental terrorism, he is dead wrong. \\ 
\cmidrule(lr){1-2}
1982 & 
\textbf{Ronald Reagan:} Contrary to some of the wild charges you may have heard, this administration has not and will not turn its back on America's elderly or America's poor. \newline
\textbf{Ronald Reagan:} Under the new budget, funding for social insurance programs will be more than double the amount spent only 6 years ago. \newline
\textbf{Ronald Reagan:} But it would be \textbf{foolish} to pretend that these or any programs cannot be made more efficient and economical. \\ 
 \bottomrule
    \end{tabular}
    \caption{[SOTU] Example usages of words from our divisive lexicon throughout SOTU addresses. In general, there are fewer uses of divisive words during SOTU speeches.}
    \label{tab:sotu_div_ex}
\end{table}

\begin{table}[]
    \centering
    \footnotesize
    \begin{tabular}{p{1cm} p{14cm}}
    \toprule
    \textbf{Year} &
    \textbf{Example Sentence(s) w/ Context} \\ 
    \cmidrule(lr){1-2}
    2020 & 
\textbf{Joe Biden:} The last thing you need is a President who ignores you, looks down at you, who just doesn't understand you. \newline
\textbf{Joe Biden:} Like President Trump. \newline
\textbf{Joe Biden:} His \textbf{reckless} personal conduct since his diagnosis, the destabilizing effect it's having on our government, is \textbf{unconscionable}. \\ 
\cmidrule(lr){1-2}
    2020 & 
\textbf{Joe Biden:} I'm so grateful to have earned the UA's endorsement — and to have 355,000 proud plumbers, pipefitters, and more behind me. \newline
\textbf{Joe Biden:} I also want to thank Rick for sharing his story with us today—and for being part of our convention this year where he shared his story with America. \newline
\textbf{Joe Biden:} Farmers all across this country have been gutted by President Trump's broken promises and \textbf{reckless} trade war. \\ 
\cmidrule(lr){1-2}
    2020 & 
\textbf{Donald Trump:} This is the most important election in the history of our country. \newline
\textbf{Donald Trump:} Six months ago I was saying, "Well, how do you compare with the last one?" \newline
\textbf{Donald Trump:} I don't know. \newline
\textbf{Donald Trump:} That was important. \newline
\textbf{Donald Trump:} The fact is, this is the single most important election in the history of our country. \newline
\textbf{Donald Trump:} And sleepy Joe Biden's made a \textbf{corrupt} bargain. \\ 
\cmidrule(lr){1-2}
2016 & 
\textbf{Donald Trump:} Now, Bernie Sanders should be angry right? \newline
\textbf{Donald Trump:} Shouldn't he be angry? \newline
\textbf{Donald Trump:} Now, I'll tell you what. \newline
\textbf{Donald Trump:} The system is rigged. \newline
\textbf{Donald Trump:} The system is rigged. \newline
\textbf{Donald Trump:} I've been saying it for a—it's rigged, and we're gonna straighten it out. \newline
\textbf{Donald Trump:} But the system is rigged. \newline
\textbf{Donald Trump:} Hillary is not the victim; the American people are the victims of this system. \newline
\textbf{Donald Trump:} So \textbf{corrupt} in so many ways. \\ 
\cmidrule(lr){1-2}
2016 & 
\textbf{Donald Trump:} I want the entire \textbf{corrupt} Washington establishment to hear and to heed the words we all will be saying right now. \newline
\textbf{Donald Trump:} When we win on November 8th, we are going to Washington, D.C. and we are going to drain the swamp. \newline
\textbf{Donald Trump:} Gonna drain the swamp. \newline
\textbf{Donald Trump:} We're gonna drain the swamp, folks. \newline
\textbf{Donald Trump:} We're gonna drain that swamp. \newline
\textbf{Donald Trump:} Another important issue for Americans is integrity in journalism. \newline
\textbf{Donald Trump:} These people are among the most \textbf{dishonest} people I've ever met, spoken to, done business with. \newline
\textbf{Donald Trump:} These are the most \textbf{dishonest} people. \\ 
\cmidrule(lr){1-2}
2016 & 
\textbf{Donald Trump:} Our trade deals, we lose \$800 billion a year on trade. \newline
\textbf{Donald Trump:} We have trade deficits. \newline
\textbf{Donald Trump:} Think of that. \newline
\textbf{Donald Trump:} Who negotiates these deals? \newline
\textbf{Donald Trump:} You know who does? \newline
\textbf{Donald Trump:} \textbf{Stupid} people. \newline
\textbf{Donald Trump:} \textbf{Stupid} people. \newline
\textbf{Donald Trump:} With very \textbf{stupid} leadership. \\ 
\cmidrule(lr){1-2}
2016 & 
\textbf{Donald Trump:} We will terminate NAFTA and get a much better deal for our workers if we can't renegotiate it properly. \newline
\textbf{Donald Trump:} We're gonna get a better deal for our workers and for our companies. \newline
\textbf{Donald Trump:} Because we cannot continue to be the people led by \textbf{stupid} people. \\ 
\cmidrule(lr){1-2}
2016 & 
\textbf{Hillary Clinton:} We should honor the men and women in uniform who fight for our country. \newline
\textbf{Hillary Clinton:} That's why I was so appalled when Donald Trump tweeted that the new effort underway to push the terrorists out of the key city of Mosul is already, and I quote him, "a total disaster" and that our country is, again a quote, "looking \textbf{dumb}." \newline
\textbf{Hillary Clinton:} Really? \newline
\textbf{Hillary Clinton:} He's declaring defeat before the battle has even started. \newline
\textbf{Hillary Clinton:} He's proving once again he is \textbf{unqualified} to be commander in chief of our military. \\ 
\cmidrule(lr){1-2}
2016 & 
\textbf{Hillary Clinton:} But this is not new. \newline
\textbf{Hillary Clinton:} I know I'm reaching out to Republicans and Independents as well as Democrats because I want to be the president for all Americans. \newline
\textbf{Hillary Clinton:} And, when you think about it, what he said at the convention, 'I alone can fix it,' runs counter to who we are as Americans. \newline
\textbf{Hillary Clinton:} We work together. \newline
\textbf{Hillary Clinton:} So there are many reasons why I think it is fair to conclude that Donald Trump is \textbf{unqualified} and unfit to be president. \\ 
\cmidrule(lr){1-2}
2008 & 
\textbf{John McCain:} In a time of trouble and danger for our country, who will put our country first? \newline
\textbf{John McCain:} In 21 months, during hundreds of speeches, town halls and debates, I have kept my promise to level with you about my plans to reform Washington and get this country moving again. \newline
\textbf{John McCain:} As a senator, I've seen the \textbf{corrupt} ways of Washington in wasteful spending and other abuses of power. \\ 
\cmidrule(lr){1-2}
2008 & 
\textbf{John McCain:} In his three short years in the Senate, he has requested nearly a billion dollars in pork projects for his state - a million dollars for every day he's been in office. \newline
\textbf{John McCain:} Far from fighting earmarks in Congress, Senator Obama has been an eager participant in this \textbf{corrupt} system. \\ 

 \bottomrule
    \end{tabular}
    \caption{[CAMPAIGN] Example usages of words from our divisive lexicon throughout campaigns. }
    \label{tab:campaign_div_ex}
\end{table}

\end{appendices}

\end{document}